\pdfoutput=1
\documentclass{article}

\usepackage{arxiv}

\usepackage[utf8]{inputenc} 
\usepackage[T1]{fontenc}    
\usepackage{lmodern}
\usepackage{hyperref}       
\usepackage{url}            
\usepackage{booktabs}       
\usepackage{amsfonts}       
\usepackage{amsmath}
\usepackage{nicefrac}       
\usepackage{microtype}      
\usepackage{graphicx}
\usepackage{doi}
\usepackage[nameinlink,sort&compress,capitalize]{cleveref}
\usepackage{adjustbox}
\usepackage{booktabs}
\usepackage{tabularx}
\usepackage{makecell}
\usepackage{xcolor}
\usepackage{xfrac}
\usepackage[numbers,sort&compress]{natbib}

\title{Bioinspired Soft Quadrotors Jointly Unlock Agility, Squeezability, and Collision Resilience}

\author{ \href{https://orcid.org/0000-0002-9904-1157}{\includegraphics[scale=0.06]{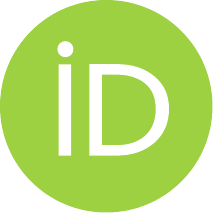}\hspace{1mm}Luca~Girardi}\thanks{Corresponding author.} \\
	Department of Environmental\\ Systems Science (D-USYS) \\
	ETH Z\"urich\\
	Z\"urich, Switzerland \\
	\href{mailto:girardil@ethz.ch}{\texttt{girardil@ethz.ch}}\\
	\And
	\href{https://orcid.org/0009-0004-7634-1538}{\includegraphics[scale=0.06]{orcid.pdf}\hspace{1mm}Gabriel~Maquignaz}\\
    Department of Environmental\\ Systems Science (D-USYS) \\
	ETH Z\"urich\\
	Z\"urich, Switzerland \\
	\href{mailto:gmaquignaz@ethz.ch}{\texttt{gmaquignaz@ethz.ch}} \\
    \And
	\href{https://orcid.org/0000-0001-6272-0212}{\includegraphics[scale=0.06]{orcid.pdf}\hspace{1mm}Stefano~Mintchev}\\
    Department of Environmental\\ Systems Science (D-USYS) \\
	ETH Z\"urich\\
	Z\"urich, Switzerland \\
	\href{mailto:smintchev@ethz.ch}{\texttt{smintchev@ethz.ch}} \\
}

\renewcommand{\shorttitle}{Bioinspired Soft Quadrotors}

\hypersetup{
pdftitle={Bioinspired Soft Quadrotors Jointly Unlock Agility, Squeezability, and Collision Resilience},
pdfsubject={cs.RO},
pdfauthor={Luca~Girardi, Gabriel~Maquignaz, Stefano~Mintchev},
pdfkeywords={aerial robotics, soft robotics, bioinspired robotics, soft aerial robotics, reconfigurable quadrotors, morphing quadrotors, collision resilient quadrotors, agile quadrotors},
}

\begin{document}
\maketitle

\begin{abstract}
	
Natural flyers use soft wings to seamlessly enable a wide range of flight behaviours, including agile manoeuvres, squeezing through narrow passageways, and withstanding collisions. In contrast, conventional quadrotor designs rely on rigid frames that support agile flight but inherently limit collision resilience and squeezability, thereby constraining flight capabilities in cluttered environments. Inspired by the anisotropic stiffness and distributed mass-energy structures observed in biological organisms, we introduce FlexiQuad, a soft-frame quadrotor design approach that limits this trade-off. We demonstrate a 405-gram FlexiQuad prototype, three orders of magnitude more compliant than conventional quadrotors, yet capable of acrobatic manoeuvres with peak speeds above 80 km/h and linear and angular accelerations exceeding 3 g and 300 rad/s\textsuperscript{2}, respectively. Analysis demonstrates it can replicate accelerations of rigid counterparts up to a thrust-to-weight ratio of 8. Simultaneously, FlexiQuad exhibits fourfold higher collision resilience, surviving frontal impacts at 5 m/s without damage and reducing destabilising forces in glancing collisions by a factor of 39. Its frame can fully compress, enabling flight through gaps as narrow as 70\% of its nominal width. Our analysis identifies an optimal structural softness range, from 0.006 to 0.77 N/mm, comparable to that of natural flyers’ wings, whereby agility, squeezability, and collision resilience are jointly achieved for FlexiQuad models from 20 to 3000 grams. FlexiQuad expands hovering drone capabilities in complex environments, enabling robust physical interactions without compromising flight performance.

\end{abstract}



\section{Introduction}

The first autonomous flight of a quadrotor in 2007 marked the dawn of the civilian drone era~\cite{gurdan2007energy,bouabdallah2007full}. In less than two decades, these vehicles have transformed industries as diverse as cinematography, geospatial mapping, precision agriculture, infrastructure inspection and emergency logistics~\cite{floreano2015science,sage2022testing}. Their flight depends on an elegantly simple architecture: a rigid, cross-shaped airframe whose four cantilevered arms carry the rotors, while a central hub houses computation, sensing and energy storage. This architecture, already pioneered at the beginning of the 20\textsuperscript{th} century by manned prototypes such as the Breguet-Richet Gyroplane No. 01~\cite{leishman2002breguet}, combines predictable dynamics with high load-bearing capacity. Augmented by modern processors and sensors, today’s rigid-frame quadrotors can perform aggressive manoeuvres, exceeding 100 km/h and sustaining accelerations multiple times that of gravity~\cite{kaufmann2023champion,foehn2022agilicious}.

However, rigid frames impose severe penalties in cluttered or confined spaces. Manoeuvring through narrow gaps favours a compact footprint, which in turn restricts payload capacity and endurance~\cite{floreano2015science,karydis2017energetics}. Additionally, rigid airframes are vulnerable to mechanical failure in the event of collisions~\cite{mintchev2017insect}. Active folding or morphing mechanisms can reduce the footprint to fly through gaps~\cite{falanga2018foldable,riviere2018agile,zhao2018design,bucki2019design,patnaik2020design}, and protective add-ons can increase collision resilience~\cite{briod2014collision,Mintchev2018bioinspired,sareh2018rotorigami}, but both approaches introduce mass and mechanical complexity that erode agility and efficiency.

\begin{figure*}[htbp]
  \centering
  \includegraphics[width=0.99\textwidth]{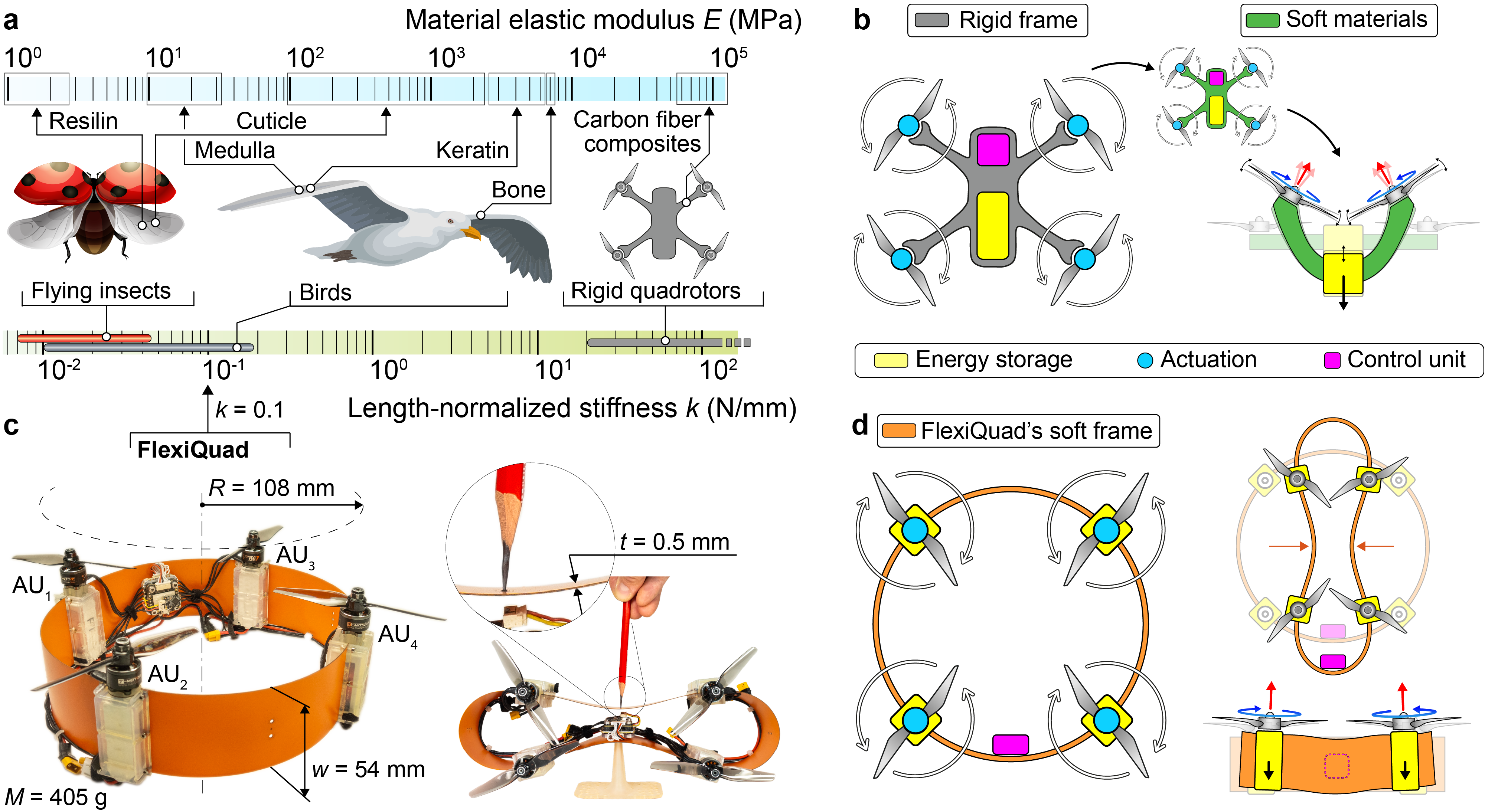}
  \caption{FlexiQuad: a bioinspired class of soft quadrotors. \textbf{a}, Natural flyers’ wings combine rigid and soft materials, resulting in structures that are orders of magnitude more flexible than the frames of conventional quadrotors. Elastic moduli ranges extracted from experimental values in the literature for bird wing bones and primary feather keratin and medulla, resilin-rich tissues in insect wing folds and tendons~\cite{appel2024resilin}, and stiffer insect wings’ chitin-rich and sclerotised cuticles\cite{vincent2004design,combes2003flexuralI}. \textbf{b}, Incorporating soft materials into the existing quadrotor frame morphology results in limited out-of-plane rigidity, leading to excessive flexural deformations that ultimately reduce efficiency and agility. \textbf{c}, Perspective and top views of a FlexiQuad model in its nominal and fully deformed configurations, with indication of its radius ($R$), frame strips width ($w$), mass ($M$), stiffness ($k$), and actuation units (AUs). Its extreme in-plane flexibility allows it to be fully compressed by a fragile pencil lead. \textbf{d}, Despite its high in-plane compliance, FlexiQuad retains sufficient out-of-plane rigidity owing to its anisotropic stiffness and decentralisation of masses beneath the rotors.}
  \label{fig:figure1}
\end{figure*}

Nature suggests a different design logic. Birds and insects rely on wings with anisotropic, compliant architectures that simultaneously deliver agility, morphing and collision resilience (\Crefsub{fig:figure1}{a}; see \Cref{sec:flyers_stiffness}). Birds’ feathers exhibit sufficient flexural stiffness to generate aerodynamic lift, while their in-plane compliance enables a continuous wing morphing~\cite{matloff2020flight}. This adaptability supports fine control over flight speed, trajectory, and stability~\cite{thomas1996flight,Lentink2007,cheney2021raptor,selim2021peregrine,harvey2022birds}, and allows birds to ‘\emph{squeeze}' and manoeuvre through confined spaces~\cite{williams2015pigeons,badger2023sideways}. Likewise, insect wings integrate rigid cuticle surfaces with soft resilin joints, enabling anisotropic deformations that enhance aerodynamic performance~\cite{young2009details,mountcastle2013wing,mistick2016wing} and dampen collisions, reducing mechanical damage and flight instability~\cite{mountcastle2014biomechanical,phan2020mechanisms}.

Soft structures have been successfully exploited in fixed~\cite{ifju2002flexible,hays2013aerodynamic,chang2020soft,sedky2024distributed} and flapping wing~\cite{ma2013controlled,ramezani2017biomimetic,karasek2018tailless} drones, but leveraging softness in quadrotors remains challenging. Replacing rigid arms with uniformly soft materials introduces undesirable flexural modes and vibrations that degrade control stability, flight efficiency and agility~\cite{ruiz2022sophie,haluvska2022soft,nguyen2023soft,dePetris2025morphy} (\Crefsub{fig:figure1}{b}).

Here we introduce FlexiQuad (\Crefsub{fig:figure1}{c}), a quadrotor class that departs from the century-old rigid-airframe design paradigm. Drawing inspiration from biological systems, FlexiQuad employs two key design principles: functional stiffness anisotropy and distributed energy mass~\cite{aubin2022towards,burden2024animals}.  This bioinspired approach endows FlexiQuad with a level of softness comparable to that found in natural flyers’ wings. Through systematic modelling and experimental validation across different weight scales (see \Cref{sec:scaling}), we demonstrate that FlexiQuad models match the agility of their rigid twins while gaining the added benefits of squeezability and collision resilience.

\section{Bioinspired Design Rationale}

FlexiQuad consists of a thin-walled, ring-shaped, soft airframe with no central body, decentralising its energy storage, actuation, and control hardware along its circumference (\Crefsub{fig:figure1}{c}; see \Cref{sec:components}). This morphology draws inspiration from biological features of functional stiffness anisotropy~\cite{liu2020structural} and distributed energy mass~\cite{aubin2022towards,burden2024animals}. 

The airframe uses a high-aspect-ratio fibreglass strip that provides compliance only in useful directions, akin to the anisotropic stiffness found in natural flyers’ wings~\cite{matloff2020flight,young2009details,mountcastle2013wing,mistick2016wing}. Specifically, the frame exhibits low resistance to in-plane compression but is $\sim$2.2 times stiffer against flexural loads aligned with rotor thrust (see \Crefsub{fig:figure1}{d} and \Cref{sec:stiffness_def}). This anisotropy ensures sufficient load-bearing capability for linear and angular accelerations during flight, while allowing the quadrotor to easily reduce its horizontal footprint for navigating confined openings and to absorb kinetic energy upon collisions.

Building on the principle of distributed energy mass, the ring-shaped airframe hosts multifunctional actuation units (\Crefsub{fig:figure1}{d}), each combining energy storage and propulsion, analogous to how vertebrate muscle metabolism provides energy from within their muscles~\cite{aubin2022towards,burden2024animals}. By decentralising the battery mass from a single central body, as in conventional quadrotors, and integrating it within the actuation units, FlexiQuad avoids the detrimental bending moments that arise when soft structures support a centrally suspended payload (\Crefsub{fig:figure1}{b}). In synergy with the frame’s anisotropy, this mass distribution ensures that the frame maintains the rotor axes in a vertical equilibrium direction, thus minimising aerodynamic lift and energy losses.

\section{Squeezability}

\begin{figure*}[htbp]
  \centering
  \includegraphics[width=0.99\textwidth]{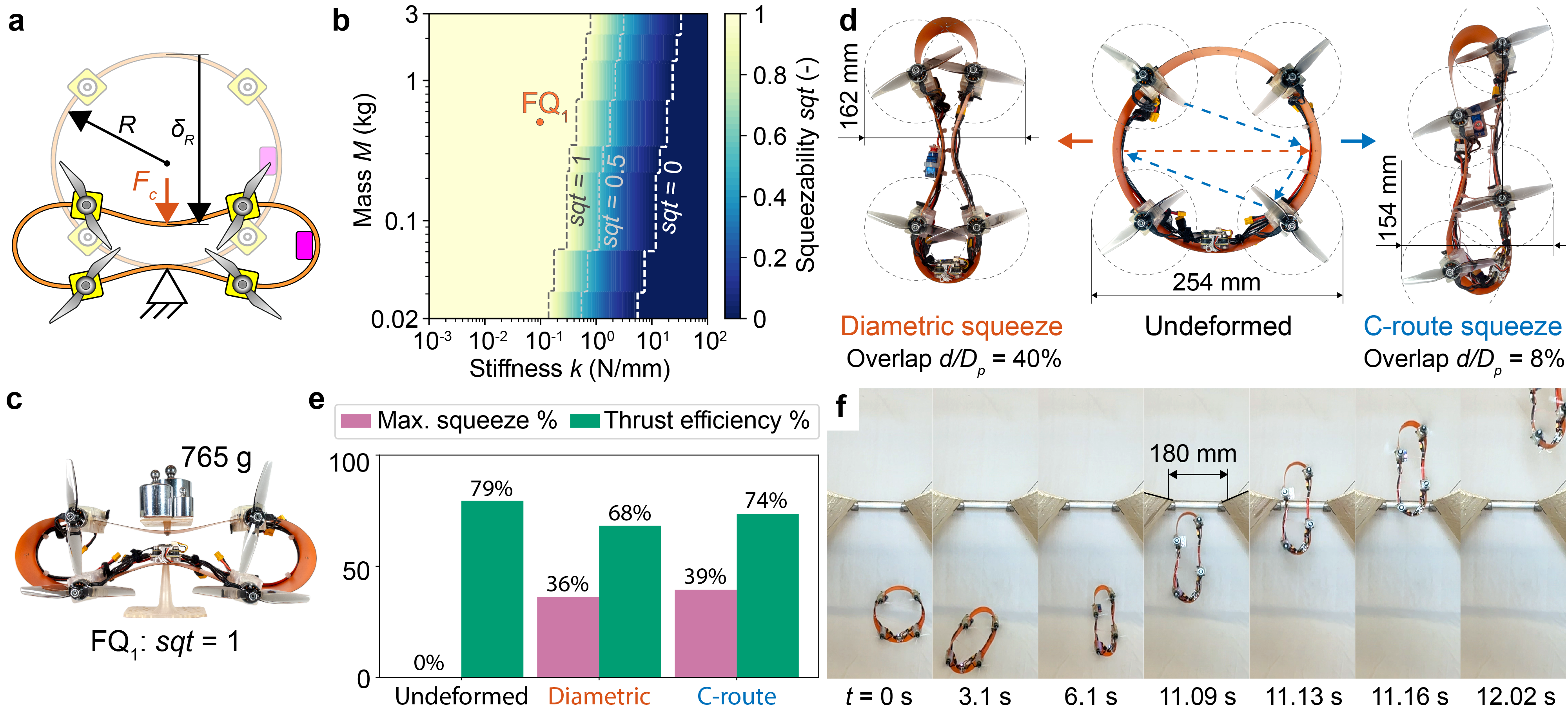}
  \caption{Squeezability. \textbf{a}, Applying a compression force ($F_C$) reduces FlexiQuad’s width by a diametric compression ($\delta_R$) in the transverse plane. \textbf{b}, Squeezability levels across varying FlexiQuad masses ($M$) and frame stiffnesses ($k$), analytically modelled by solving the planar Euler elastica. \textbf{c}, FlexiQuad model ($M$ = 0.405 kg, $k$ = 0.1 N/mm) subject to a 90\% frontal squeeze under weights of mass
  $\sim1.9 M$. \textbf{d}, A single actuator driving a string-and-pulley system enables frame squeezing. Alternative string-routing configurations leading to distinct equilibrium morphologies with variable propeller overlap distances ($d$), relative to the propeller diameter ($D_p$). \textbf{e}, Squeezing is accompanied by a decrease in energy efficiency as propellers are drawn close together and relative overlap $d/D_p$ increases. \textbf{f}, Active squeezing enables flight through gaps 0.7 as narrow as the drone’s nominal width (254 mm).}
  \label{fig:figure2}
\end{figure*}

Many birds navigate narrow openings in vegetation by tucking their wings, momentarily sacrificing flight efficiency to reduce their effective size~\cite{williams2015pigeons,badger2023sideways}. Inspired by these wing-tucking strategies, we designed FlexiQuad with a soft airframe capable of compressing into compact shapes (\Crefsub{fig:figure2}{a}), a property we term \emph{squeezability}. This enables low-energy morphological adaptation, achievable actively via low-strength actuation or passively under external forces.

We define the squeezability index ($sqt$) as the maximum achievable diametric compression ($\delta_{R,max}$), normalised by the initial airframe diameter 2$R$ (see \Crefsub{fig:figure2}{a}, and \Cref{sec:sqt_model}):
\begin{equation}
    sqt = \frac{\delta_{R,max}}{2R}
    \label{eq:sqt}
\end{equation}
Using an analytical beam model, we quantified $sqt$ across stiffness and mass variations within FlexiQuad’s parametric design space (\Crefsub{fig:figure2}{b} and \Cref{sec:beam_model}). Softer frames, identified by the region $k\leq0.55M^{0.35}$ (log-transformed linear regression, $R_{\mathrm{fit}}^2$ = 0.955), all allow complete squeezability ($sqt$ = 1). In contrast, stiffer frames exceed safe material stress thresholds before achieving full compression, limiting $sqt$ to values below 1. Within this region, equal $sqt$ levels follow an approximate scaling power law of $k \propto M^{0.35}$, testifying that squeezability scales favourably with FlexiQuad’s mass. Consequently, heavier and larger drones, which stand to benefit the most from reducing their overall size for flight in cluttered environments, exhibit higher squeezability for a given frame-equivalent stiffness.

Within the region of complete squeezability, the required compression force $F_C$ (in newtons) to fully squeeze ($sqt$ = 1) scales approximately as:
\begin{equation}
    F_C|_{sqt=1} \approx 8.67\: k\: M^{0.35}
    \label{eq:sq_force}
\end{equation}
with stiffness $k$ in N/mm, and drone mass $M$ in kg (see \Crefmultisubfiglist{fig:figure2}{a}{c}, \Crefmultisubfiglist{fig:figure7}{c}{d}). Leveraging linear scaling of $F_C|_{sqt=1}$ with frame stiffness $k$, soft FlexiQuad variants can achieve active squeezing using a low-strength, single-actuator string-and-pulley mechanism. By altering the routing of the string, FlexiQuad implements different squeezing modes that trade off flight efficiency for navigational capability in confined spaces. Each squeezing mode achieves a static equilibrium configuration determined by the string’s routing geometry and spool length (\Crefsub{fig:figure2}{d}, Supplementary Video~1). Diametric squeezing (\Crefsub{fig:figure2}{d}, left) preserves symmetry, but overlapping propellers limit the minimum width to 162 mm and reduce thrust efficiency from 79\% (undeformed) to 68\% (\Crefsub{fig:figure2}{e}). A C-routing scheme (\Crefsub{fig:figure2}{d}, right) mitigates these limitations by alternately positioning propellers along FlexiQuad’s length, achieving a narrower width of 154 mm and improving thrust efficiency to 73\% (see \Cref{sec:sqt_model}).

We experimentally demonstrated active squeezing using a 405 g FlexiQuad model ($k$ = 0.1 N/mm, $sqt$ = 1) actuated by a 9-g servomotor and pulley system. Employing C-route squeezing, FlexiQuad successfully navigated through a 180 mm wide gap, 70\% of its original width, while maintaining stable and controlled flight (\Crefsub{fig:figure2}{f}, Supplementary Video~2). It also demonstrated squeeze-to-grasp functionality by grasping, transporting, and releasing in-flight a 136 g first-aid kit without requiring installation of additional grasping mechanisms (Supplementary Video~2). 

\section{Collision Resilience}

Collision resilience defines an aerial system’s ability to endure collisions with obstacles while preserving its structural integrity and operational functionality. Insects provide a compelling model for this capability: through biomechanical adaptations such as flexible joints and wing-folding, they avoid cumulative damage~\cite{mountcastle2014biomechanical}, and dampen collision forces, allowing rapid recovery of flight stability~\cite{phan2020mechanisms}. By contrast, rigid quadrotors lack this resilience, as full-body horizontal collisions result in impulsive energy dissipation into the airframe, often causing structural or electronic failure~\cite{Mintchev2018bioinspired}. 

\begin{figure*}[htbp]
  \centering
  \includegraphics[width=0.99\textwidth]{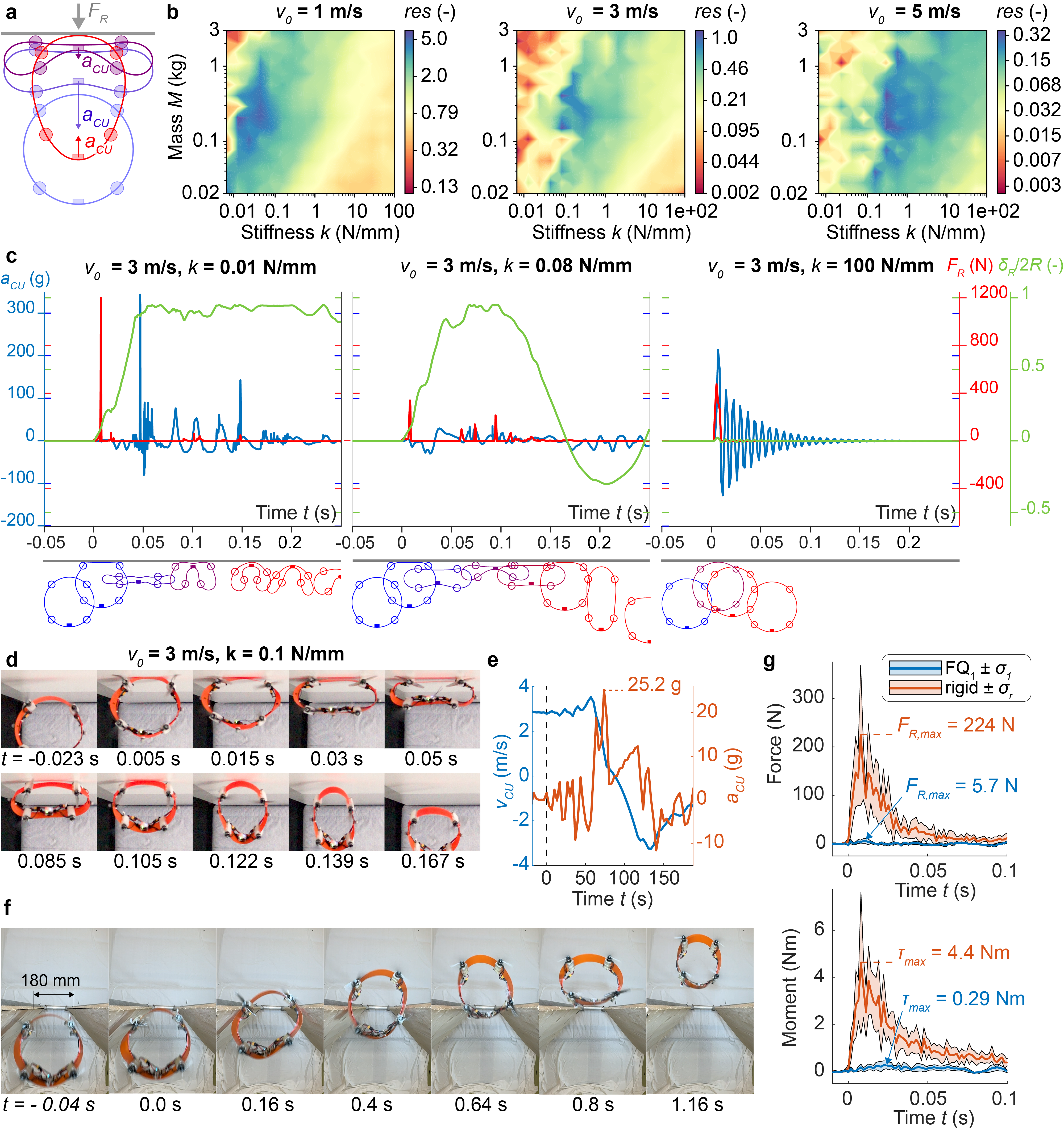}
  \vspace{-7pt}\caption{Collision resilience. \textbf{a}, During frontal collisions, FlexiQuad’s compliant frame preferentially deforms within its transverse plane. The control unit (CU) experiences a deceleration $a_{CU}$, and the impact plane exerts a reaction collision force $F_R$. \textbf{b}, Heatmaps of the collision resilience metric ($res$) across varying FlexiQuad masses ($M$) and frame stiffnesses ($k$) at impact velocities $v_0$ = 1, 3, and 5 m/s, computed via finite element analysis. \textbf{c}, Temporal profiles of deceleration, collision force, and frontal relative diametric compression ($\delta_R/2R$) for frames with increasing stiffness at a collision velocity of 3 m/s. \textbf{d}, Experimental recording of a frontal collision of a FlexiQuad model ($M$ = 0.405 kg, $k$ = 0.1 N/mm) at 3 m/s. \textbf{e}, Time-dependent velocity ($v_{CU}$) and acceleration ($a_{CU}$) profiles at the CU measured during the frontal collision with a motion capture system. \textbf{f}, Sequential snapshots of a \emph{squeeze-and-fly} manoeuvre, showing momentum-driven passive squeeze through a narrow gap (180 mm), 70\% as narrow as the model’s nominal width. \textbf{g}, Experimental temporal profiles of force and torque measured during a simulated glancing collision on the FlexiQuad model FQ\textsubscript{1} ($M$ = 0.405 kg, $k$ = 0.1 N/mm) and on a rigid quadrotor with identical mass and avionics. Solid lines represent the mean, and shaded regions denote $\pm$1 standard deviation ($\sigma$; n = 6 and 8 trials for soft and rigid, respectively).}
  \label{fig:figure3}
\end{figure*}

FlexiQuad addresses these shortcomings by deforming upon impact, thus absorbing kinetic energy through elastic deflections over a duration 10 times longer than that of a rigid drone (\Crefmultisubfiglist{fig:figure3}{a}{c}). Moreover, its decentralised mass distribution leads to lower peak forces applied asynchronously across the rigid components, in contrast to the single, high-impulse peak force observed in rigid frames.

Because FlexiQuad’s collisions fragment into multiple sub-impacts, its primary vulnerability to damage is governed by the peak acceleration experienced by the control unit (CU), the most impact-sensitive component. We therefore define \emph{collision resilience} ($res$) as the inverse of the peak acceleration (in g) measured at the control unit during frontal collisions at velocity ($v$), normalised by 50 g (see \Cref{sec:collision_model}): 
\begin{equation}
    res|_{v} = \biggl( \frac{a_{CU,max}|_v}{50} \biggr)^{-1}
    \label{eq:res}
\end{equation}

Through finite element analysis, we evaluated FlexiQuad’s collision resilience across a range of frame stiffness and drone mass values at impact velocities of 1, 3, and 5 m/s (\Crefsub{fig:figure3}{b}). Overly soft frames fail to absorb energy effectively, resulting in the sequential impact of the rigid internal components (\Crefsub{fig:figure3}{c}, left). Rigid frames produce a single short-duration impact with small compression and high peak deceleration (\Crefsub{fig:figure3}{c}, right). Optimal frame stiffness allows instead FlexiQuad to dissipate energy gradually along the full deformation, prolonging impact duration and reducing peak deceleration (\Crefsub{fig:figure3}{c}, centre). Therefore, the analysis reveals regions of optimal resilience that shift toward higher stiffness as collision velocity increases, indicating that higher speeds require stiffer frames to absorb and redirect the greater kinetic energy before full compression occurs (\Crefsub{fig:figure3}{b}).

Our FlexiQuad prototype withstood high-speed ($v_0$ = 3 and 4.5 m/s) frontal collisions without incurring structural or functional damage, reducing peak accelerations transmitted to onboard electronics, measuring only 25.2 g during a collision at 3 m/s (\Crefmultisubfiglist{fig:figure3}{d}{e}, Supplementary Video~3). Furthermore, we demonstrated FlexiQuad leverages synergies between squeezability and resilience to passively compress to traverse gaps as narrow as 70\% of its nominal width, exploiting momentum-driven deformation to achieve passage approximately 10 times faster than the previously described controlled ‘\emph{squeeze-and-fly}' approach and without the need for additional actuators for active squeezing control (\Crefsub{fig:figure3}{f}, Supplementary Video~4). 

Beyond full-body impacts, glancing collisions involving rotor arms or propellers are common during high-speed navigation in cluttered environments. While such interactions may not immediately damage the structure of rigid drones, they often result in mission failure due to destabilisation or propeller breakage (Supplementary Video~5). FlexiQuad’s compliance reduces peak loads by up to 39-fold during these events (\Crefsub{fig:figure3}{g}), substantially containing destabilising load disturbances and the risk of propeller damage. This compliance also acts as a form of mechanical intelligence, passively dampening destabilising moments on the frame. In our experiments, this effect reduced destabilising torques by a factor of 15, offering potential for more robust post-collision recovery and control.

\section{Agility}

Agility is a critical evolutionary trait in birds, enabling rapid changes in speed and direction to evade predators, capture prey, or avoid collisions~\cite{moore2015outrun,chin2022birds}. Similarly, agility enables quadrotors to perform rapid, dynamic manoeuvres essential for navigating cluttered environments at high speeds, known to extend operational range~\cite{bauersfeld2022range}.

Multirotor agility refers to the drone’s capability to rapidly change linear velocity and body rates while remaining within stable flight conditions~\cite{kumar2012opportunities}, aligned with its definition for birds~\cite{harvey2022birds,dudley2002mechanisms,dakin2018morphology}. To evaluate FlexiQuad’s agility, we estimated its maximum linear and angular accelerations across different M and k configurations, using a parametric FE model to simulate dynamic free-body responses under increasing acceleration commands (Supplementary Video~6). Observing instances of structural collapse, self-contact of the frame, or inter-propeller contact determined the upper bounds to the admissible accelerations for each configuration. 

\begin{figure*}[htbp]
  \centering
  \includegraphics[width=0.99\textwidth]{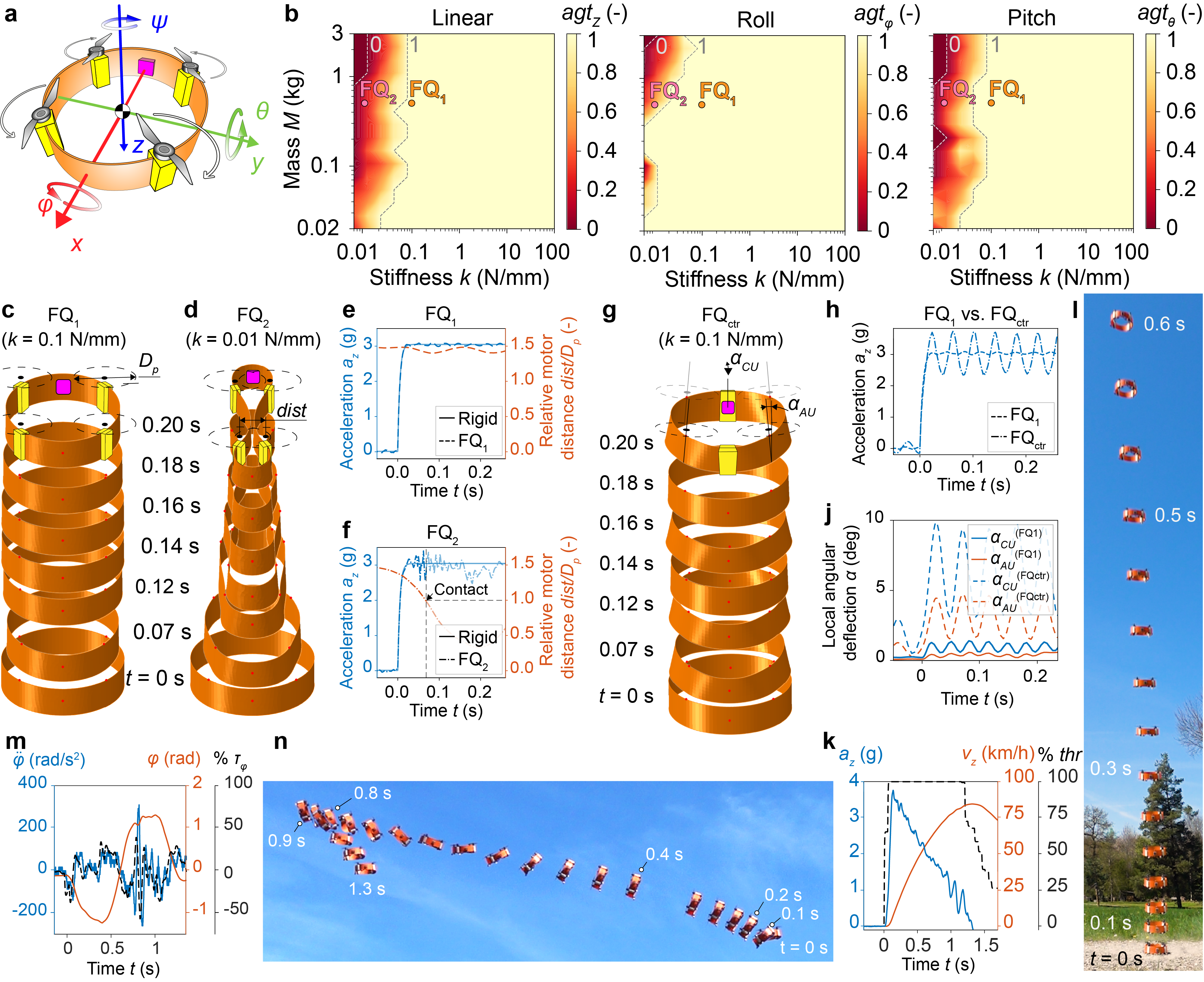}
  \caption{Agility. \textbf{a}, Schematics of FlexiQuad’s body coordinate frame ($x$, $y$, $z$) and the corresponding roll ($\phi$), pitch ($\vartheta$), and yaw ($\psi$) angles. \textbf{b}, Linear, roll, and pitch agility results in FlexiQuad’s ($M$, $k$) design space, with indication of 0.405-kg FlexiQuad models FQ\textsubscript{1} ($k$ = 0.1 N/mm) and FQ\textsubscript{2} ($k$ = 0.01 N/mm). \textbf{c},\textbf{d}, Simulation results of step linear acceleration command for 0.405-kg FQ\textsubscript{1} (\textbf{c}) and FQ\textsubscript{2} (\textbf{d}) at TWR = 4. \textbf{e},\textbf{f}, Comparison of vertical centre-of-mass accelerations ($a_z$) and lateral actuator inter-axis distance ($dist/D_p$) \textendash\enspace relative to propeller diameter $D_p$ \textendash\enspace between FQ\textsubscript{1} (\textbf{e}) and FQ\textsubscript{2} (\textbf{f}) from simulation results in (\textbf{c}), (\textbf{d}). \textbf{g}, Simulation results of a step linear acceleration of an equivalent 0.405-kg model, FQ\textsubscript{ctr}, with concentrated battery mass (indicated in yellow) and $k$ = 0.1 N/mm. \textbf{h}, Comparison of vertical acceleration $a_z$ between FQ\textsubscript{1} and FQ\textsubscript{ctr} from simulations in (\textbf{c}) and (\textbf{g}). \textbf{j},  Comparison of angular deviation from the vertical at the control unit ($\alpha_{CU}$) and at the actuation units ($\alpha_{AU}$, averaged across all units) between FQ\textsubscript{1} and FQ\textsubscript{ctr} from simulations in (\textbf{c}) and (\textbf{g}). \textbf{k},\textbf{l}, Snapshots and graphs of vertical acceleration ($a_z$) and velocity ($v_z$) measured during experimental maximum step vertical acceleration with throttle percentage indicated as \% $thr$, using a FlexiQuad model ($M$ = 0.405 kg, $k$ = 0.1 N/mm, TWR = 4.15). \textbf{m},\textbf{n}, Snapshots and graphs of roll acceleration ($\ddot \phi$, blue) and angle ($\phi$, red) measured during aggressive lateral dodging manoeuvre with percent reference roll torque command (\% $\tau_\phi$, dashed black), using the FlexiQuad model FQ\textsubscript{1} ($M$ = 0.405 kg, $k$ = 0.1 N/mm), TWR = 4.15.}
  \label{fig:figure4}
\end{figure*}

From the simulated accelerations, we determined agility indices ($agt$) for a selected manoeuvre as the ratio of FlexiQuad’s maximum admissible acceleration from hover to that of a rigid twin ($k$ = 100 N/mm) of identical inertial properties and TWR = 8:
\begin{equation}
    agt_{\zeta} = \frac{\textrm{max}(\ddot \zeta_{\textrm{FQ})}}{\textrm{max}(\ddot \zeta_{\textrm{rigid}})},  \qquad \zeta =z,\varphi,\vartheta,
    \label{eq:agt}
\end{equation}
with $\zeta$ indicating linear Z-axis position ($\zeta$ = $z$, global coordinates), and body roll ($\varphi$), and pitch ($\vartheta$) angles (\Crefsub{fig:figure4}{a}), while $\ddot \zeta$ indicates double time differentiation. Yaw ($\psi$) manoeuvres were analysed separately as agile trajectories don’t require aggressive yaw accelerations (see \Cref{sec:agility_model}).

For each manoeuvre, FlexiQuad’s agility remained unaffected by airframe stiffness across a vast region of the ($M$, $k$) design space (\Crefsub{fig:figure4}{b}). Here, the deformations resulting from the manoeuvre’s dynamics were sufficiently small to minimally alter the relative distance and orientation of actuation units (AUs) and match rigid quadrotors’ acceleration levels (\Crefmultisubfiglist{fig:figure4}{c}{e}, \Crefmultisubfigrange{fig:figure11}{a}{d}). In contrast, configurations with excessively low stiffness (e.g. $k$ < 0.2 N/mm, for $M$ = 0.405 kg, TWR = 4) experienced structural compression leading to inter-propeller contact at identical acceleration inputs, constraining dynamic response to lower agility values (\Crefmultisubfiglist{fig:figure4}{d}{f}, \Crefmultisubfigrange{fig:figure11}{a}{d}, Supplementary Video~6). 

Distributing energy storage at the AUs was pivotal to achieving rigid-like agility at lower stiffness. In contrast, quadrotor designs with centralised battery mass and identical k exhibited oscillations 13 and 8 times larger at the AUs and CU under acceleration, respectively (\Crefmultisubfigrange{fig:figure4}{g}{j}). When adding closed-loop control, these oscillations would appear as disturbances read by the inertial measurement unit at the CU, necessitating aggressive filtering to maintain flight stability. However, since the dominant oscillation frequency overlaps with the range of agile manoeuvres (\Crefsub{fig:figure11}{e}), filtering reduces control bandwidth, inherently trading off agility for stability. 

Flight experiments with a FlexiQuad model ($M$ = 405 g, $k$ = 0.1 N/mm, TWR = 4.15) demonstrated peak linear accelerations exceeding 3 g (\Crefmultisubfiglist{fig:figure4}{k}{l}), peak angular accelerations above 300 rad/s\textsuperscript{2} (\Crefmultisubfiglist{fig:figure4}{m}{n}), and vertical and horizontal speed surpassing 84 and 77 km/h, respectively (\Crefsub{fig:figure4}{k}, \Crefsub{fig:figure11}{e}, Supplementary Video~7), exceeding the optimal cruise speed reported for efficient flight in small quadrotors~\cite{bauersfeld2022range} (see \Cref{sec:agility_model}).

\section{Discussion}

Through a systematic exploration of the soft-quadrotor design space, we demonstrated that FlexiQuad airframes with stiffness values more than three orders of magnitude lower than conventional quadrotors can still sustain the high instantaneous linear and angular accelerations required for agile flight, while enabling the ability to reduce their effective size to traverse narrow gaps and withstand high-speed collisions. In contrast to existing soft-quadrotor design attempts~\cite{ruiz2022sophie,haluvska2022soft,nguyen2023soft,dePetris2025morphy}, this substantial compliance is achieved through a circular frame composed of high-aspect-ratio, thin shells of rigid materials arranged strategically to form a functionally anisotropic structure. This configuration allows preserving the typical load-bearing and control advantages of rigid quadrotors while simultaneously introducing the collision resilience and morphing capabilities observed in soft quadrotors. Furthermore, decentralising the energy sources by positioning them axially under the actuators proved essential for minimising destabilising oscillations and out-of-plane deformations that would otherwise constrain agility. 

\begin{figure}[htbp]
  \centering
  \includegraphics[width=0.99\textwidth]{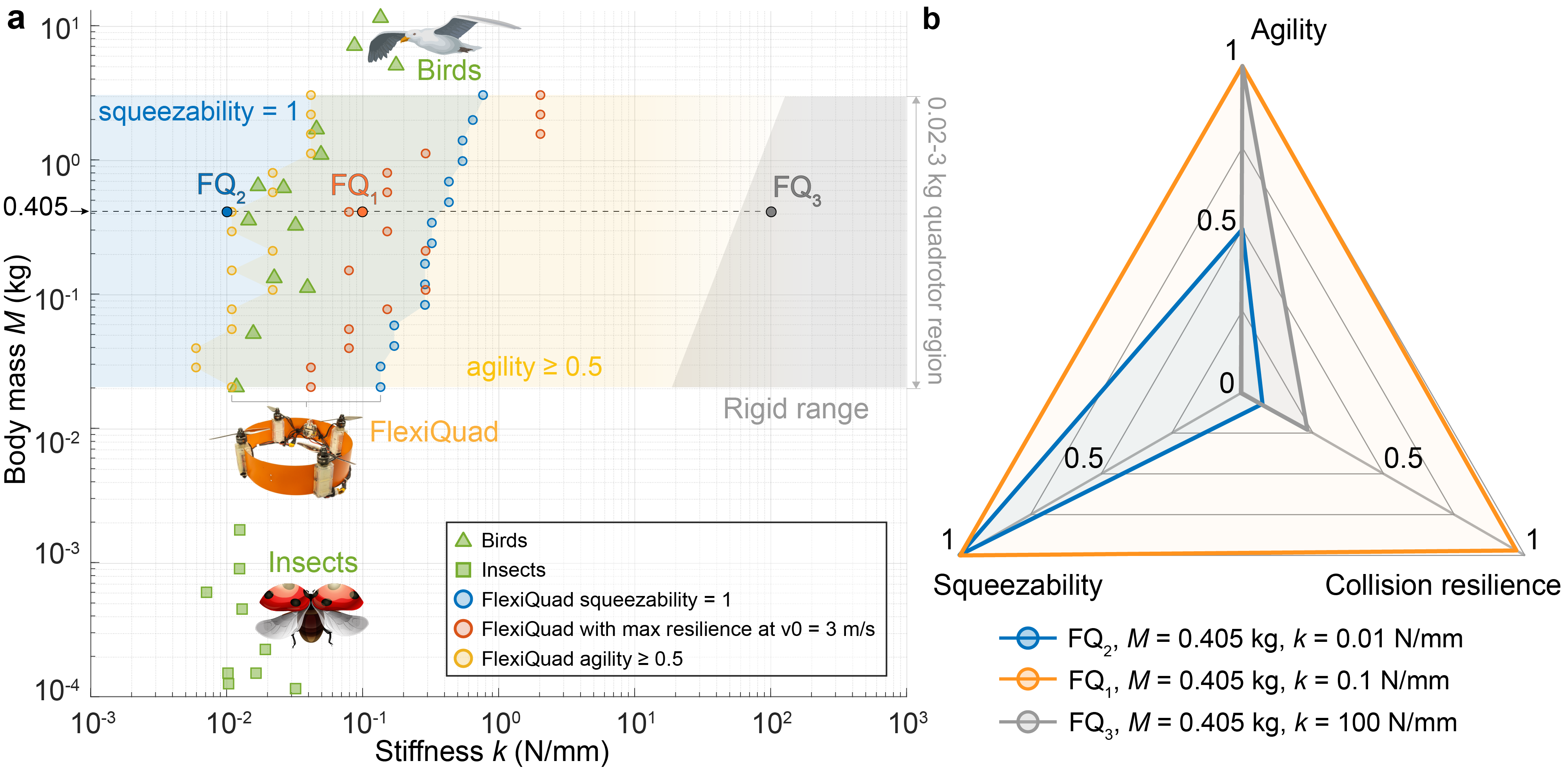}
  \caption{Stiffness and performance comparison. \textbf{a}, Comparison in the stiffness-mass plane between FlexiQuad, biological flyers, and rigid quadrotors. Colour-shaded regions indicate FlexiQuad’s parameter combinations guaranteeing complete squeezability ($sqt$ = 1, blue) and agility greater than 0.5 (yellow). Blue points indicate the upper bound of $k$ at different $M$ for squeezability = 1, yellow points the lower bound for agility $\geq$ 0.5 (corresponding to TWR = 4.5), red points the optimal $k$ for collision resilience at $v_0$ = 3 m/s. Grey shading indicates mass and stiffness parameter values found in commercially available rigid quadrotor frames. \textbf{b}, Performance comparison between 0.405-kg FlexiQuad models with varying stiffness: one that is too soft (FQ\textsubscript{2}, $k$ = 0.01 N/mm), one in the optimal stiffness range (FQ\textsubscript{1}, $k$ = 0.1 N/mm), and a full rigid one (FQ\textsubscript{3}, $k$ = 100 N/mm).}
  \label{fig:figure5}
\end{figure}

Notably, it is possible to identify an optimal stiffness region from 0.006 to 0.77 N/mm that closely overlaps with stiffness values measured in the wings of biological flyers, ranging from sub-gram insects to birds weighing more than 10 kilograms (\Crefsub{fig:figure5}{a}). FlexiQuad designs within this range achieve agile flight with thrust-to-weight ratios $\geq$4.5 (agility $\geq$0.5), exceeding those of typical survey and inspection quadrotors (TWR $\approx$ 2) and aligning with high-performance autonomous racing drones (TWR $\approx$ 4-5~\cite{foehn2022agilicious}). At the same time, this stiffness interval enables complete airframe compression and limits peak collision decelerations to less than one-quarter of those recorded in rigid frames. Outside this optimal range, Flexiquad models become either too stiff (FQ\textsubscript{3}) or too soft (FQ\textsubscript{2}), resulting in trade-offs between agility, squeezability, and collision resilience (\Crefsub{fig:figure5}{b}).

The successful integration of biological-level compliance into hovering drones now shifts the challenge toward perception and control. Real-time shape sensing capable of reconstructing the airframe’s instantaneous deformation field~\cite{shih2020electronic} would enable morphology-aware controllers and adaptive filtering that suppress oscillations during collisions or aggressive manoeuvres, thereby supporting robust autonomous flight~\cite{dePetris2025morphy}.  Embracing softness may ultimately require a re-examination of conventional operational paradigms, enabling the purposeful use of behaviours once considered undesirable in rigid quadrotors. For instance, the frame’s intrinsic oscillations could provide proprioceptive~\cite{weber2021wing,deCroon2022accommodating} or exteroceptive~\cite{deCroon2021enhancing} cues, and navigation algorithms might exploit squeezability and collision resilience to execute deliberate impacts that produce rapid re-orientations~\cite{jayaram2018transition} or to access complex, cluttered environments~\cite{fabris2022crash}. Because soft machines tightly couple deformation, sensing and actuation, future progress will demand the co-optimisation of morphology and control policy~\cite{howard2019evolving,gupta2021embodied}. Bio-inspired evolutionary algorithms that iteratively refine both morphology and controller offer a principled route through this expanded design space, promising task-specific “\emph{body-brain}” solutions. 

Despite these challenges, FlexiQuad stands as a compelling proof of concept for soft aerial robotics. By uniting rigid-like flight agility with the resilience and squeezability offered by compliant structures, soft drones will transform interactive tasks ranging from inspection of ageing infrastructure to autonomous search and data collection in confined, cluttered natural or man-made environments.


\section{Methods}

\subsection{FlexiQuad Components and Fabrication}
\label{sec:components}

\begin{figure*}[htbp]
  \centering
  \includegraphics[width=0.99\textwidth]{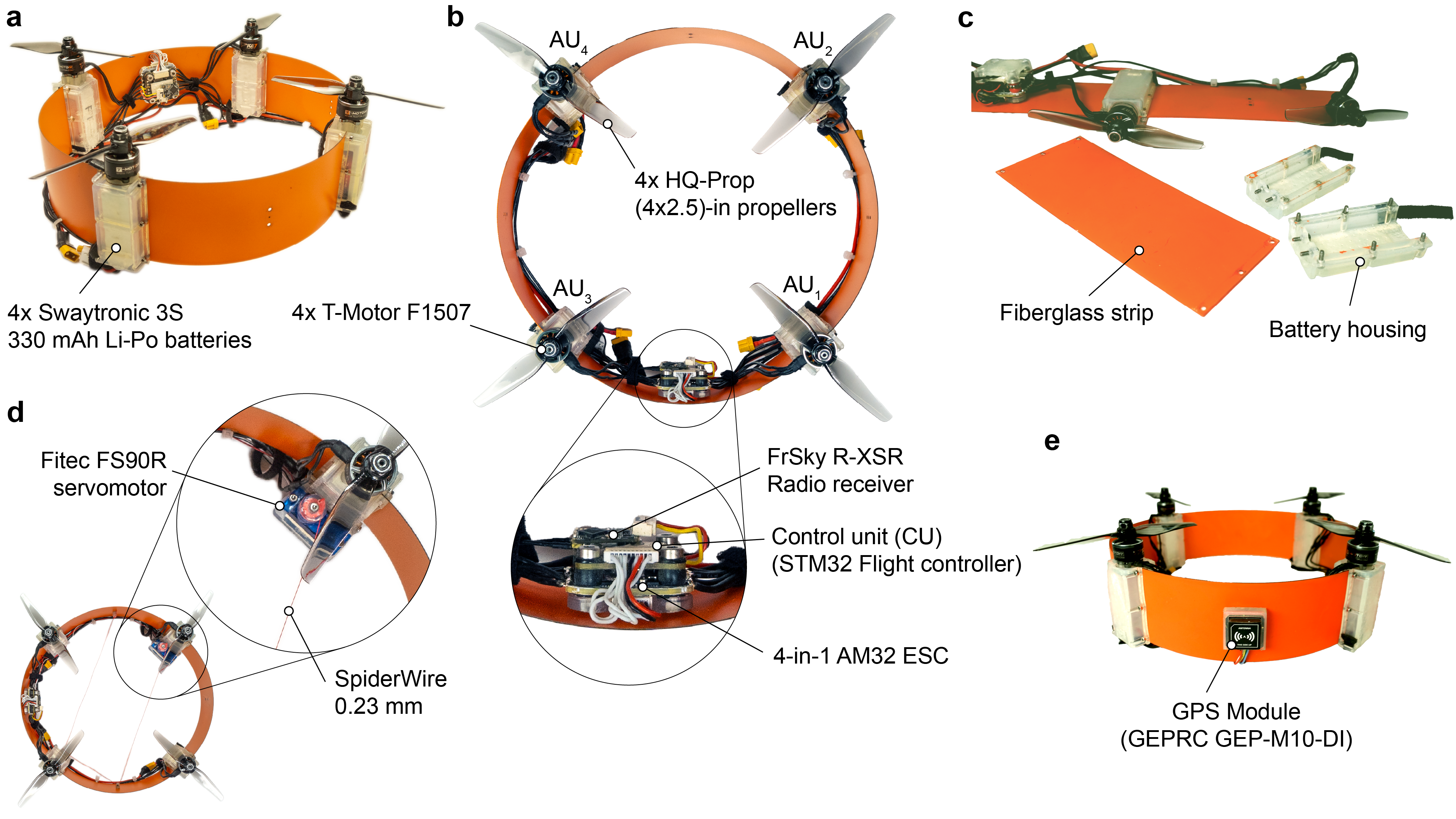}
  \caption{FlexiQuad’s components. \textbf{a}, Perspective view of FlexiQuad FQ\textsubscript{1} model. \textbf{b}, Top view of Flexiquad with a zoom-in on the control unit (CU). \textbf{c}, Partially disassembled view of the FQ\textsubscript{1} model, highlighting the flat fibreglass strip segments and the two shells composing all resin-3D printed battery housings. \textbf{d}, Top view of FlexiQuad with a zoom-in on the servomotor responsible for actively controlling frame squeeze. \textbf{e}, Rear perspective view of the FlexiQuad model used in the outdoor agility and speed experiments, highlighting the addition of the GPS module, rigidly attached behind the CU.}
  \label{fig:figure6}
\end{figure*}

The FlexiQuad model used in our demonstrations has a total takeoff mass of 405 g and a diameter of 216 mm. Its modular frame consists of four thin glass-fibre composite strips (Swiss Composites, FR-4 HF woven), each measuring (145.5 × 54 × 0.5) mm, laser-cut using a CO\textsubscript{2} laser (Trotec Speedy 360, 60 W) aligned with the weaving pattern of the composite laminate. These strips are fastened with M2 screws to four actuation units (AU), forming a closed-loop airframe (\Crefmultisubfigrange{fig:figure6}{a}{c}). Each AU, whose frame is printed with high-toughness resin (Formlabs Durable Resin) using an LCD-stereolithography 3D printer (Formlabs Form 4), serves as housing for one battery and as a rigid mount for one propulsion unit. Each propulsion unit comprises a brushless motor (T-Motor F1507/KV3800) paired with polycarbonate two-blade 4-inch propellers (HQ-Prop 4x2.5). The power system consists of four 330 mAh three-cell Li-Po batteries (Swaytronic, 3S, 330 mAh, 60C) connected in parallel to an integrated 4-in-1 electronic speed controller (Flywoo GOKU G45M 45 A Mini ESC 2-6S AM32). The ESC is stacked with the control unit (CU, flight controller MatekSys F405-miniTE) and secured to the rearmost frame strip using a custom 3D-printed mount. FlexiQuad was remotely controlled via a radio transmitter (FrSky Taranis X9 Lite) linked to an onboard receiver (FrSky R-XSR) communicating with the flight controller via serial protocol (SBUS). Electrical wiring for signal transmission and power distribution is routed along the frame and secured with zip ties and Velcro strips. 

For the in-flight active squeezing demonstrations, we mounted a continuous-rotation servomotor (Fitec FS90R) on the frame at different locations (\Crefmultisubfiglist{fig:figure2}{e}{f}, and \refsub{fig:figure6}{d}), which drove the spooling of a high-strength, braided, poly-ethylene string (SpiderWire Stealth Smooth 0.23 mm) through 3D-printed pulleys and anchor points at designated locations. Servomotor commands were given via the radio controller, and a flight controller’s servo output drove the motor actuation. For the outdoor flight tests, a GPS and barometer module (GEPRC GEP-M10-DI) was mounted behind the CU stack to track position and velocity (\Crefsub{fig:figure6}{e}).

\subsection{Biological Flyers Wing Stiffness and Mass Scaling}
\label{sec:flyers_stiffness}

Propulsive efficiency likely underlies the convergent evolution of powered winged flight in birds, insects, and bats~\cite{taylor2003flying}. Across taxa, scales, and fluid media, both flying and swimming animals consistently exhibit compliant bending patterns of their propulsors, which contribute to enhanced propulsion efficiency~\cite{combes2001shape,lucas2014bending}. 

The stiffness of these propulsors is commonly measured as mean flexural stiffness ($EI$), representing the local resistance of a cross-section to bending moments under the assumption of an equivalent uniform cross-section~\cite{combes2003flexuralI,worcester1996scaling}. Values of $EI$ are experimentally determined in bird primary feathers~\cite{worcester1996scaling} and insect wings~\cite{combes2003flexuralI} using two-point static bending tests and applying the linearised cantilevered beam model:
\begin{equation}
    EI = \frac{FL^3}{3\delta}
    \label{eq:cantilever}
\end{equation}
where $F$ is the applied force, $L$ is the cantilever distance from the loading point to the anchor, and $\delta$ is the corresponding deflection.

In insects, allometric studies show a strong correlation between flexural stiffness and wing length ($L_w$), with $EI\propto L_w^{2.97}$. A mass-to-wing-length scaling factor of 0.33~\cite{greenewalt1975flight,pennycuick2008allometry}, further implies a nearly linear relationship between $EI$ and body mass. In birds, scaling studies indicate that feather flexural stiffness scales with mass to an exponent between 1.08 and 1.24, corresponding to exponents of 3.03 to 3.76 when expressed in terms of wing length~\cite{worcester1996scaling}.

To compare the flying animal wing response to loading across scales, we define the normalised stiffness $k$ as the ratio of applied force $F$ to the resulting displacement $\delta$, analogous to a linearised spring constant:
\begin{equation}
    k = \frac{F}{\delta}
    \label{eq:k_definition}
\end{equation}
By combining \Cref{eq:cantilever,eq:k_definition}, $k$ can be estimated from values of $EI$ in the literature as:
\begin{equation}
    k = 3\frac{EI}{L^3}
    \label{eq:k_function}
\end{equation}
Because EI scales approximately as $L_w^3$ in both insects and birds, and $L\propto L_w$~\cite{combes2003flexuralI,sullivan2019scaling}, $k$ provides a measure of stiffness that is less sensitive to size, enabling comparison across different scales.

Within bird wings' load-bearing regions, the lowest values of $EI$ are measured in primary bird feathers~\cite{worcester1996scaling,bachmann2012flexural,wang2012size}. For this reason, we estimated birds’ values of $k$ in \Crefsub{fig:figure1}{a} by applying \Cref{eq:k_function} to the experimental primary feathers $EI$ data from 13 bird species ranging from 0.02 to 11.3 kg of body mass~\cite{worcester1996scaling}. In that study, values of $EI$  were measured via two-point bending of feathers attached at the base of the calamus and loaded at a cantilever length $L$ corresponding to the length of the bare calamus plus the proximal-most fourth of the vane.

For insects, we applied \Cref{eq:k_function} to experimental values of spanwise flexural stiffness $EI$ measured in 16 flying insect species from six orders, ranging masses from 10\textsuperscript{-5} to 0.012 kg~\cite{combes2003flexuralI}. In that study, values of $EI$ were measured with two-point bending tests on wings attached at the base and loaded at $L$ = 0.7 $\times L_w$, where $L_w$ is the wing length.

\subsection{FlexiQuad Airframe Stiffness Definition}
\label{sec:stiffness_def}

\begin{figure*}[htbp]
  \centering
  \includegraphics[width=0.97\textwidth]{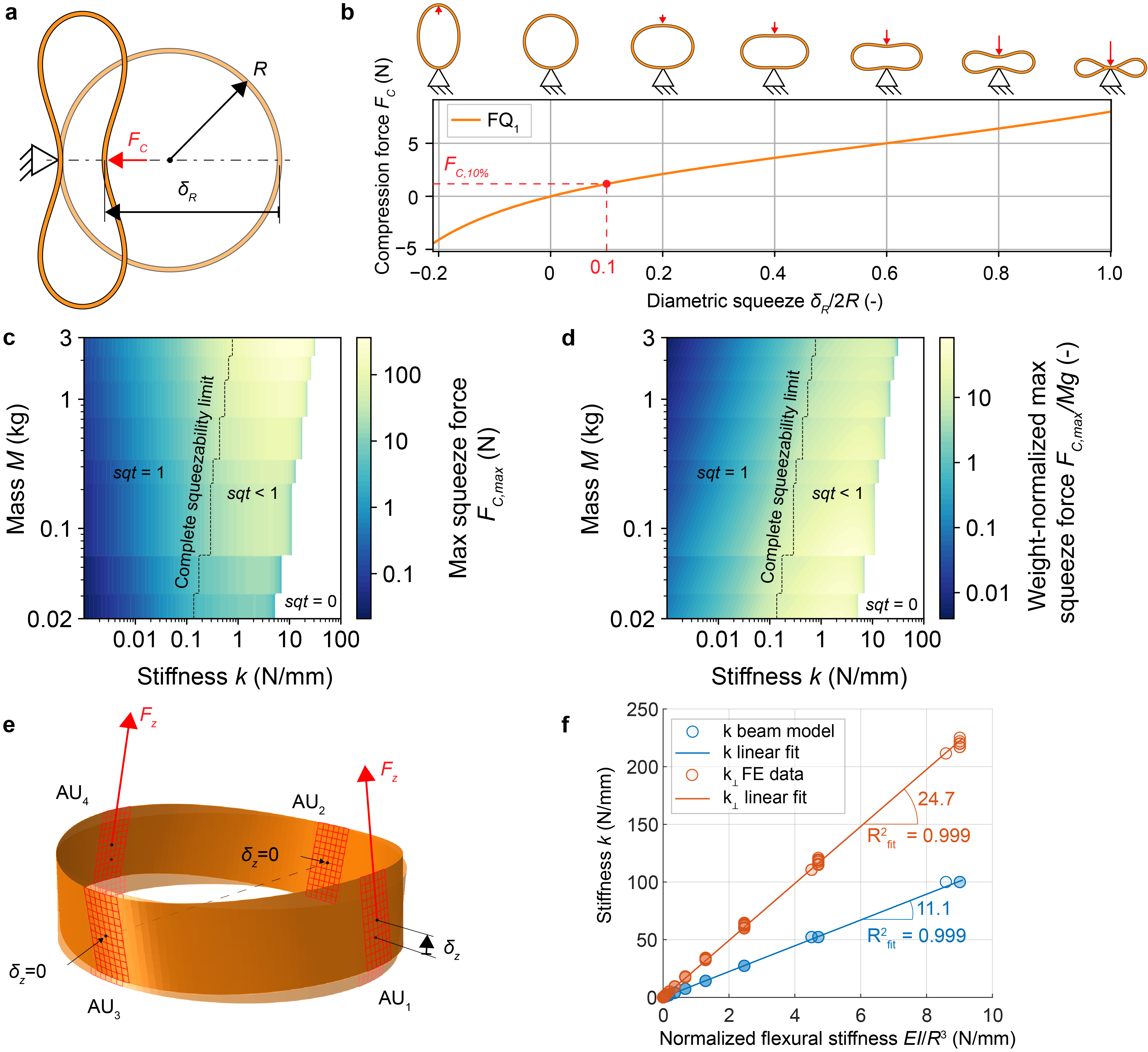}
  \vspace{-5pt}\caption{FlexiQuad airframe stiffness definition. \textbf{a}, In-plane stiffness ($k$) definition. Schematic representation of diametric compression (\emph{squeeze}, $\delta_R$) within the transverse plane as an effect of the compression force ($F_C$). \textbf{b}, Graph of $F_C$ as a function of relative squeeze ($\delta_R/2R$) for the FlexiQuad model FQ\textsubscript{1}, computed with the Euler-elastica model, highlighting the compression force at 10\% squeeze ($F_{C,10\%}$) used in the frame stiffness definition (\Cref{eq:k_definition_FQ}). \textbf{c},\textbf{d}, Heatmaps of the maximum squeezing force ($F_{C,max}=F_C |_{sqt=1}$, \textbf{c}) and its weight-normalised values ($F_{C,max}/Mg$, \textbf{d}) necessary to achieve the maximum admissible squeeze across variations of FlexiQuad stiffness ($k$) and mass ($M$). The regions of complete, incomplete, and null squeezability ($sqt$ = 1, $sqt$ < 1, and $sqt$ = 0, respectively, with approximate limit for $sqt$ = 1 as $k\leq 0.55\: M^{0.35}$) are indicated. All values are obtained from the Euler-elastica model. \textbf{e}, Out-of-plane stiffness ($k_\perp$) definition. Overlapped view of the deformed (solid) and undeformed (partially transparent) solutions of the finite element (FE) out-of-plane bending analysis. Actuation units (AUs) AU\textsubscript{2} and AU\textsubscript{3} are constrained within the transverse plane, AU\textsubscript{1} and AU\textsubscript{4} are displaced out-of-plane by $\delta_z$, producing reaction forces $F_z$. \textbf{f}, Graphs of linear scaling laws for $k$ and $k_\perp$ proportionally to FlexiQuad’s constitutive and geometrical properties $EI/R^3$. Data points from the Euler beam model and FE analysis for $k$ and $k_\perp$, respectively. Both linear regressions with zero intercept and coefficients of determination $R_{fit}^2 \geq 0.999$.}
  \label{fig:figure7}
\end{figure*}

We define FlexiQuad’s airframe stiffness ($k$) as the ratio between the diametrical compression force ($F_C$) required to reduce the ring width by 10\% of its original diameter (2$R$), and the resulting radial displacement ($\delta_R$; \Crefmultisubfiglist{fig:figure7}{a}{b}), consistent with the stiffness definition used for biological flyers (\Cref{eq:k_definition}):
\begin{equation}
    k = \frac{F_C}{\delta_R}\biggr|_{\delta_R/2R=0.1}
    \label{eq:k_definition_FQ}
\end{equation}

FlexiQuad’s flexural stiffness ($EI$) depends on the material and geometry of the fibreglass ring. The effective flexural modulus ($E$) corresponds to the longitudinal modulus ($E_{11}$) of the orthotropic laminate. The orthotropic formulation applies to single-ply, balanced woven laminates such as the selected FR-4 glass-fibre composite. The cubic dependence of $EI$ on length ($EI\propto L^3$) observed in flying animals (\Cref{eq:cantilever}) also applies to FlexiQuad’s radius $R$:
\begin{equation}
    k = \tilde k\frac{EI}{R^3}
    \label{eq:k_function_FQ}
\end{equation}
Here, $\tilde k$ is a non-dimensional stiffness constant, which arises solely from the problem’s geometry and remains independent of material stiffness or dimensions. To determine $\tilde k$, we solved the non-dimensionalised Euler elastica for an initially circular beam representing FlexiQuad's fibreglass strip and loading condition, applying \Cref{eq:k_function_FQ} to the dimensionless form:
\begin{equation}
    \tilde k = \frac{\tilde F_C}{\tilde \delta_R}\biggr|_{\tilde \delta_R=0.1} = 11.1
    \label{eq:k_tilde}
\end{equation}
where non-dimensional variables are defined as $\tilde F_C = F_C R^2/EI$, $\tilde \delta_R = \delta_R/2R$. We validated \Cref{eq:k_function_FQ,eq:k_tilde} using finite-element (FE) simulations across the full ($M$, $k$) design space (see \Cref{sec:beam_model}).

Whereas $k$ quantifies the in-plane stiffness, functional for squeezing and resilience, out-of-plane stiffness ($k_\perp$) governs resistance to thrust impulses, roll and pitch torque, and aerodynamic drag on the airframe. Analogous to $k$, we define $k_\perp$ as the vertical force ($F_z$) required at AUs 1 and 4 (diametrically opposite) to induce an out-of-plane displacement $\delta_z$ = 0.1$R$, while AUs 2 and 3 (lying within the other diagonal) are constrained within the transverse plane (\Crefsub{fig:figure7}{e}):
\begin{equation}
    k_\perp = \frac{F_z^{(AU_1)}}{\delta_z^{(AU_1)}}\biggr|_{\delta_z^{(AU_1)}/R=0.1} = 11.1
    \label{eq:k_perp_definition}
\end{equation}
While $k$ results primarily from in-plane bending, $k_\perp$ reflects a combination of bending and twisting effects when deforming the strip out-of-plane. Nevertheless, an approximate non-dimensional form analogous to \Cref{eq:k_function_FQ} also holds:
\begin{equation}
    k_\perp \approx \tilde k_\perp\frac{EI}{R^3}
    \label{eq:k_perp_function_FQ}
\end{equation}
where $\tilde k_\perp$ = 24.7 was estimated by linear regression between $EI/R^3$ and $k_\perp$ across 16x16 log-spaced combinations in the ($M$, $k$) design space (\Cref{tab:table1}, see \Cref{sec:scaling}), with coefficient of determination $R^2_{fit}$ = 0.999 (\Crefsub{fig:figure7}{f}). This approximation assumes a balanced weave in the composite laminate and thin plate condition (strips' width-to-thickness ratio $\geq$10).

Combining \Cref{eq:k_function_FQ,eq:k_perp_function_FQ}, we identified a stiffness anisotropy index $k_\perp/k \approx $ 2.23, indicating how much more compliant FlexiQuad is in-plane than out-of-plane. This designable anisotropy enables directions of higher compliance for squeezing and resisting collisions, while preserving thrust-bearing stiffness in other directions.

To study the scaling of $k_\perp$, verify \Cref{eq:k_perp_function_FQ}, and determine $\tilde k_\perp$, we developed a 3D finite-element model of FlexiQuad in Abaqus 2024 (Dassault Systèmes). The ring-shaped airframe was modelled using S4R shell elements with rectangular cross-sections matching the fibreglass strips and lamina material properties. The influence of rigid components \textendash\enspace including batteries, PUs, and CU \textendash\enspace was represented by kinematic couplings between control points at their airframe locations and corresponding connection regions. Boundary conditions were applied at the battery centres of AUs 2 and 3, and vertical displacements $u_3$ = $\delta_z$ = 0.1$R$ were imposed at propeller centres of AUs 1 and 4 (\Crefsub{fig:figure7}{e}). Simulations used the static solver in Abaqus/Standard. We evaluated the 256 ($M$, $k$) parameter combinations, computed $k_\perp$ with \Cref{eq:k_perp_definition} using reaction forces at AUs 2 and 3, and validated the approximation in \Cref{eq:k_perp_function_FQ} with the regression discussed above.

\subsection{FlexiQuad Scaling Model}
\label{sec:scaling}

FlexiQuad is designed to cover take-off masses $M$ from 0.02 to 3 kg, ranging from the smallest autonomous quadrotors~\cite{giernacki2017crazyflie,muller2021funfiiber} to the upper limit of small-scale hovering robots reported in the literature~\cite{karydis2017energetics}. The airframe stiffness $k$ ranges from 0.006 to 100 N/mm, encompassing the flexibility of the most compliant animal wings (see \Cref{sec:flyers_stiffness}) to the rigidity of carbon fibre frames.

\begin{figure*}[htbp]
  \centering
  \includegraphics[width=0.99\textwidth]{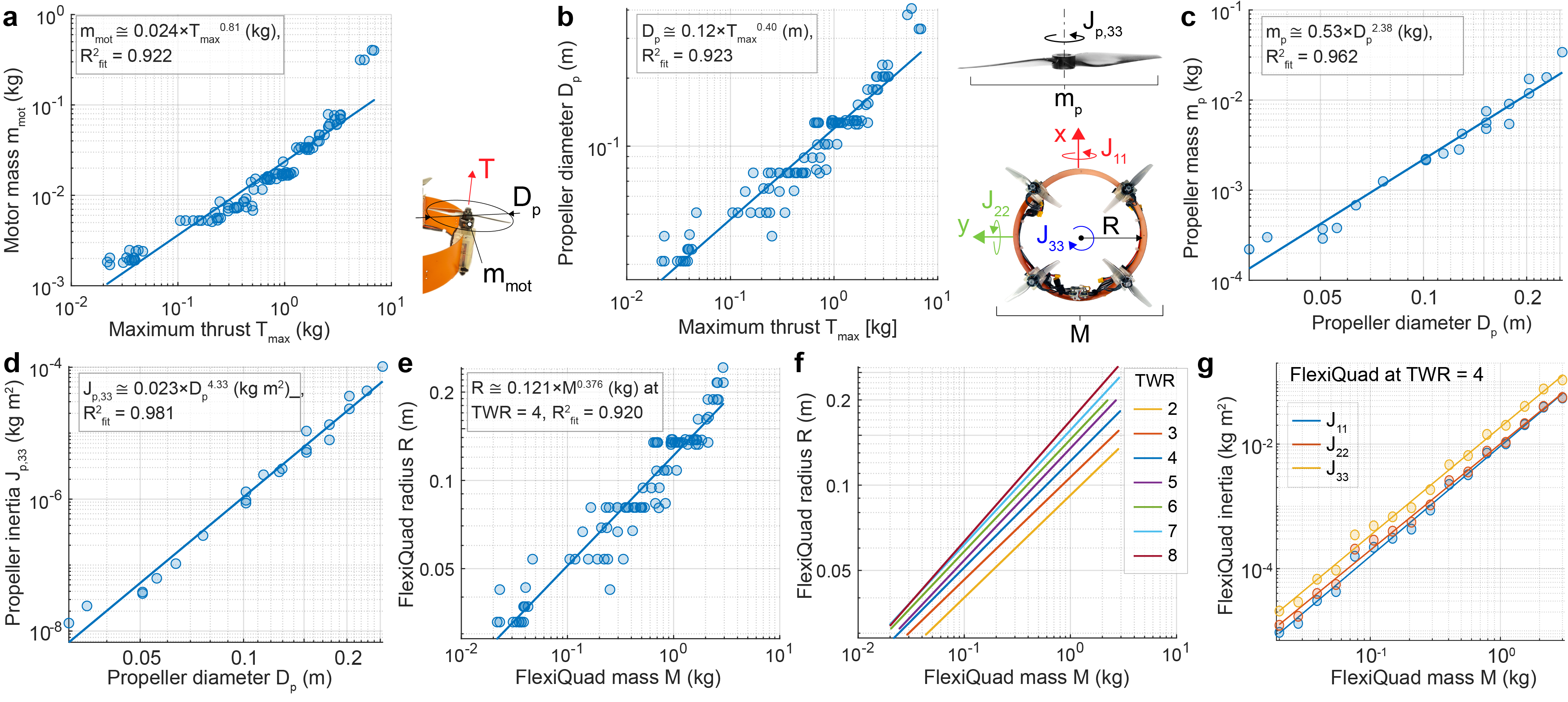}
  \caption{Empirical size and inertial scaling models. \textbf{a},\textbf{b}, Motor mass ($m_{mot}$, \textbf{a}) and highest-thrust propeller diameter ($D_p$, \textbf{b}) versus maximum produced thrust ($T_{max}$) for multirotor brushless-DC motors (BLDCMs); linear regressions in the log-transformed space (n = 105) with data points extracted from commercially available multirotor BLDCM specifications. \textbf{c},\textbf{d}, Propeller mass ($m_p$, \textbf{c}) and principal inertia about its axis ($J_{p,33}$, \textbf{d}) versus $D_p$; linear regressions in the log-transformed space (n = 22) with mass data points from commercially available specifications and inertia values estimated from 3D CAD models. \textbf{e},\textbf{f}, FlexiQuad radius ($R$) versus take-off mass ($M$) at TWR = 4 (\textbf{e}) and with varying TWR = 2, 3, …, 8 (\textbf{f}). Data points from our scaling model were evaluated on parameter combinations reported in \Cref{tab:table1}; linear regression in the log-transformed space (n = 256 for each TWR). \textbf{g}, FlexiQuad moments of inertia about body axes X ($J_{11}$), Y ($J_{22}$), and Z ($J_{33}$) versus take off mass ($M$) at TWR = 4. Data points from inertia calculation on the finite-element (FE) model; linear regression in the log-transformed space (n = 256 for each TWR value). }
  \label{fig:figure8}
\end{figure*}

Commercial multirotors within this scale range typically employ brushless direct-current motors (BLDCMs). BLDCMs and the attached propellers form propulsion units (PU) capable of generating thrust-to-weight ratios (TWRs) from two (e.g., lightweight camera drones) to five or more (e.g., first-person view (FPV) racing drones~\cite{foehn2022agilicious}). We studied 105 commercially available BLDCMs mounting their highest-thrust propeller to determine an empirical power scaling law between maximum thrust ($T_{max}$) produced by these PUs and their size and inertial properties (\Crefmultisubfigrange{fig:figure8}{a}{d}): 
\begin{equation}
    \begin{split}
        m_{mot} \approx 0.024\:T_{max}^{0.81},\qquad R^2_{fit} = 0.922,\\
        D_p \approx 0.12\:T_{max}^{0.40},\qquad R^2_{fit} = 0.923,\\
        m_p \approx 0.53\:D_p^{2.38},\qquad R^2_{fit} = 0.962,\\
        J_{p,33} \approx 0.023\:D_p^{4.33},\qquad R^2_{fit} = 0.981\\
    \end{split}
    \label{eq:scaling}
\end{equation}
Here, $m_{mot}$ and $m_p$ are the BLDCM and propeller masses, respectively, $D_p$ is the propeller diameter, $J_{p,33}$ the propeller’s principal moment of inertia about its spinning axis; all regressions are in SI units. In line with the analysed design space, we only analysed PUs with $T_{max}$ in the 0.01-6 range.
\begin{table*}[htbp]
  \small
  \centering
  \begin{tabular}{|c|c|}
    \hline
    \thead{\textbf{Mass}\\$M$ (kg)} & \thead{\textbf{Stiffness}\\$k$ (N/mm)} \\
    \hline\hline
    0.020 & 0.006 \\
    0.028 & 0.011 \\
    0.039 & 0.022 \\
    0.054 & 0.042 \\
    0.076 & 0.080 \\
    0.106 & 0.153 \\
    0.148 & 0.293 \\
    0.207 & 0.560 \\
    0.289 & 1.07  \\
    0.404 & 2.08  \\
    0.565 & 3.92  \\
    0.789 & 7.49  \\
    1.10  & 14.3  \\
    1.54  & 27.4  \\
    2.15  & 52.3  \\
    3.00  & 100   \\
    \hline
  \end{tabular}
  \caption{Parametric design space. Selected values of masses $M$ and stiffnesses $k$ for the ($M$, $k$) parametric design space used in the FE models for the collision resilience and agility analyses. All 256 possible ($M$, $k$) combinations were simulated. }
  \label{tab:table1}
\end{table*}

In FlexiQuad’s design, the inter-axis distance between adjacent PUs is set to 1.5 times the propeller diameter ($D_p$), ensuring thrust losses due to inter-propeller aerodynamic interaction remain below 1\%~\cite{lee2020rotor}. Consequently, $D_p$ determines FlexiQuad’s wheelbase, and thus the airframe’s radius $R$, as follows: 
\begin{equation}
    R = 1.5\:D_p\frac{\sqrt{2}}{2} \approx 1.06\:D_p
    \label{eq:radius_function}
\end{equation}

We set the width $w$ of the fibreglass strips to $w$ = $R$/2, thus fixing the aspect ratio between FlexiQuad’s airframe radius $R$ and its height ($w$) to two. This way, any combination of FlexiQuad’s mass and thrust requirements uniquely determines its size within the design space, resulting in the following mass-scaling power laws for FlexiQuad’s radius and principal moments of inertia ($J_{ii}$, \Crefmultisubfigrange{fig:figure8}{e}{g}): 
\begin{equation}
    \begin{split}
        R \approx 0.121\:M^{0.376},\qquad R^2_{fit} = 0.920,\\
        J_{11} \approx (9.24\times 10^{-3})\:M^{1.77},\qquad R^2_{fit} = 0.995,\\
        J_{22} \approx (1.01\times 10^{-2})\:M^{1.71},\qquad R^2_{fit} = 0.994,\\    
        J_{33} \approx (1.86\times 10^{-2})\:M^{1.74},\qquad R^2_{fit} = 0.995,\\
    \end{split}
    \label{eq:R_J_scaling}
\end{equation}
Here \textendash\enspace and for all analyses except agility \textendash\enspace we fixed TWR = 4, a common value in agile quadrotors~\cite{foehn2022agilicious}. Whereas, when studying FlexiQuad’s agility, we explore values of TWR from 2 to 8, which result in slight variations of size scaling laws (\Crefsub{fig:figure8}{f}). 

At any given mass ($M$), selecting FlexiQuad’s k determines the necessary flexural stiffness EI from \Cref{eq:k_function_FQ}:
\begin{equation}
    EI = \frac{kR^3}{\tilde k}
    \label{eq:EI_FQ}
\end{equation}
From the relationship of the second moment of area of the rectangular cross-section ($I$) and by selecting the material’s flexural modulus ($E$), we determine the thickness ($t$) of the laminate strip composing the airframe as follows:
\begin{equation}
    t = \sqrt[3]{\frac{12}{Ew}\frac{kR^3}{\tilde k}} = 2\sqrt[3]{3\frac{kR^2}{E\tilde k}}
    \label{eq:thickness}
\end{equation}
Therefore, any combination of $M$ and $k$ in the design space completely determines FlexiQuad’s inertial and geometric quantities.

\subsection{Squeezability Model and Experiments}
\label{sec:sqt_model}

We name \emph{squeeze} FlexiQuad’s diametric compression $\delta_R$. FlexiQuad achieves its maximum squeeze ($\delta_{R,max}$), used in the definition of squeezability ($sqt$) in \Cref{eq:sqt}, either when two diametrically opposite points in the frame touch ($\delta_{R,max}$ = 2$R$) or when bending stresses in the strip laminate, reach the ultimate strength of the material ($\sigma_R$), scaled by a safety factor of 1.5:
\begin{equation}
    \delta_{R,max} = \mathrm{min}\bigl (2R,\delta_R \bigr|_{\sigma_{max}=\sfrac{\sigma_R}{1.5}}\bigr )
    \label{eq:max_squeeze}
\end{equation}
where $\sigma_{max}$ is the maximum bending stress in the material at squeeze $\delta_R$, and 1.5 is a common safety factor for non-critical components for aviation. To determine both $sqt$ and the corresponding diametric compression force $F_C|_{\delta_R=\delta_{R,max}}$ for each combination of FlexiQuad’s $M$ and $k$ design parameters, we modelled the airframe analytically as a planar beam (see \Cref{sec:beam_model}). The model approximates non-dimensional compression force $\tilde F_C = F_C R^2/EI$ and curvature $\tilde \kappa = 2R\kappa$ at all points in the frame as a function of the non-dimensional radial compression $\tilde \delta_R = \delta_R/2R$. Thus, dimensional values for different design parameter combinations follow from geometric and constitutive properties with multiplicative scaling: 
\begin{equation}
    \begin{split}
        F_C = \frac{EI}{R^2 \tilde F_C},\\
        \kappa = \frac{\tilde \kappa}{2R}
    \end{split}
    \label{eq:redimensionalization}
\end{equation}

The point of maximum curvature $\kappa_{max}(\tilde \delta_R)$ experiences the highest material stress $\sigma_{max}(\tilde\delta_R)$ for any $\tilde\delta_R>0$, where $\sigma_{max}$ is the equivalent von Mises stress in the simplified pure bending case ($\sigma_{max}=|\sigma_{11}|$, with 1 being the longitudinal strip director). Assuming a linear stress-strain relationship of the laminate material for $\sigma_{max}\leq\sigma_R/1.5$, $\sigma_{max}$ follows from the maximum local change in curvature along the fibreglass strips ($|\kappa_{max}(\tilde \delta_R) - \kappa_0|$) from the initial curvature of the strips ($\kappa_0$) during squeezing:
\begin{equation}
    \sigma_{max}(\tilde \delta_R) = |\kappa_{max}(\tilde \delta_R) - \kappa_0|\frac{Et}{2}
    \label{eq:max_stress}
\end{equation}
with $t$ being the laminate strip thickness. If the strips are obtained from a flat plate, then $\kappa_0$ = 0, as in the FlexiQuad model FQ\textsubscript{1}, whereas if the strips were laminated with the desired initial curvature of radius $R$, then $\kappa_0 = 1/R$. 

Because the strip thickness $t$ relates to FlexiQuad’s stiffness $k$ as $t\propto \sqrt[3]{k}$ (see \Cref{eq:thickness}), stiffer FlexiQuad models reach the material breakdown at smaller curvatures and are more prone to break for $\tilde \delta_R<1$. Using \Cref{eq:max_squeeze,eq:max_stress}, we computed $\delta_{R,max}$ within FlexiQuad’s design space. Through \Cref{eq:redimensionalization}, we computed the maximum compression force ($F_{C,max}$) for each parameter combination, defined as the compression force corresponding to maximal squeeze: $F_{C,max} = F_C |_{\delta_R=\delta_{R,max}}$.

The approximate relation in \Cref{eq:sq_force} is obtained by studying compression forces only in the region where complete squeeze is achievable ($\tilde\delta_{R,max} = 1 \Rightarrow sqt = 1$) via multivariate linear regression in the log-transformed space between independent variables $M$ and $k$ and dependent variable $F_C|_{sqt=1}$. The regression’s coefficient of determination was $R_{fit}^2$ = 0.997 in the transformed space. See \Cref{sec:beam_model} for further details on the beam model and its validation using a 3D numerical model.

We determined thrust efficiency values shown in \Crefsub{fig:figure2}{e} by measuring the electrical power consumed by each FlexiQuad configuration at hover conditions and normalising it by the baseline power consumption of four isolated PUs operating without aerodynamic interference. We secured the FlexiQuad model in its undeformed, diametric-squeeze, and C-squeeze configurations to a six-axis force-torque sensor (Rokubi, Bota Systems) using custom laser-cut brackets that maintained the intended geometry. For each configuration, we recorded electrical power drawn from a regulated DC power supply (PSI 9040-40 T 640 W, Elektro-Automatik GmbH) at a constant voltage of 12.4 V when achieving a collective thrust of 4.0 N, as measured by the force-torque sensor. The reference power was obtained by measuring a single PU rigidly attached to the sensor, generating 1.0 N of thrust under identical conditions, and multiplying the result by four.

\subsection{Planar Non-Dimensional Beam Model}
\label{sec:beam_model}

We modelled FlexiQuad’s ring-shaped airframe as two specular, semi-circular, dimensionless, inextensible planar beams with controlled endpoints and uniform rectangular cross-section. The extremities of each beam lie within the ring diameter and control deformation through their relative displacement. The relationship between normalised squeeze ($\tilde \delta_R$) and the dimensionless compressive force applied to the ring ($\tilde F_C$) follows from elliptic integral solutions to the Euler elastica differential equation~\cite{byrd2013handbook}. Although $\tilde F_C (\tilde \delta_R)$ cannot be expressed in closed form, the functions $\tilde \delta_R(\eta)$ and $\tilde F_C(\eta)$ can be formulated explicitly in terms of the elliptic modulus $\eta$~\cite{cazzolli2019snapping}. 

\begin{figure*}[htbp]
  \centering
  \includegraphics[width=0.99\textwidth]{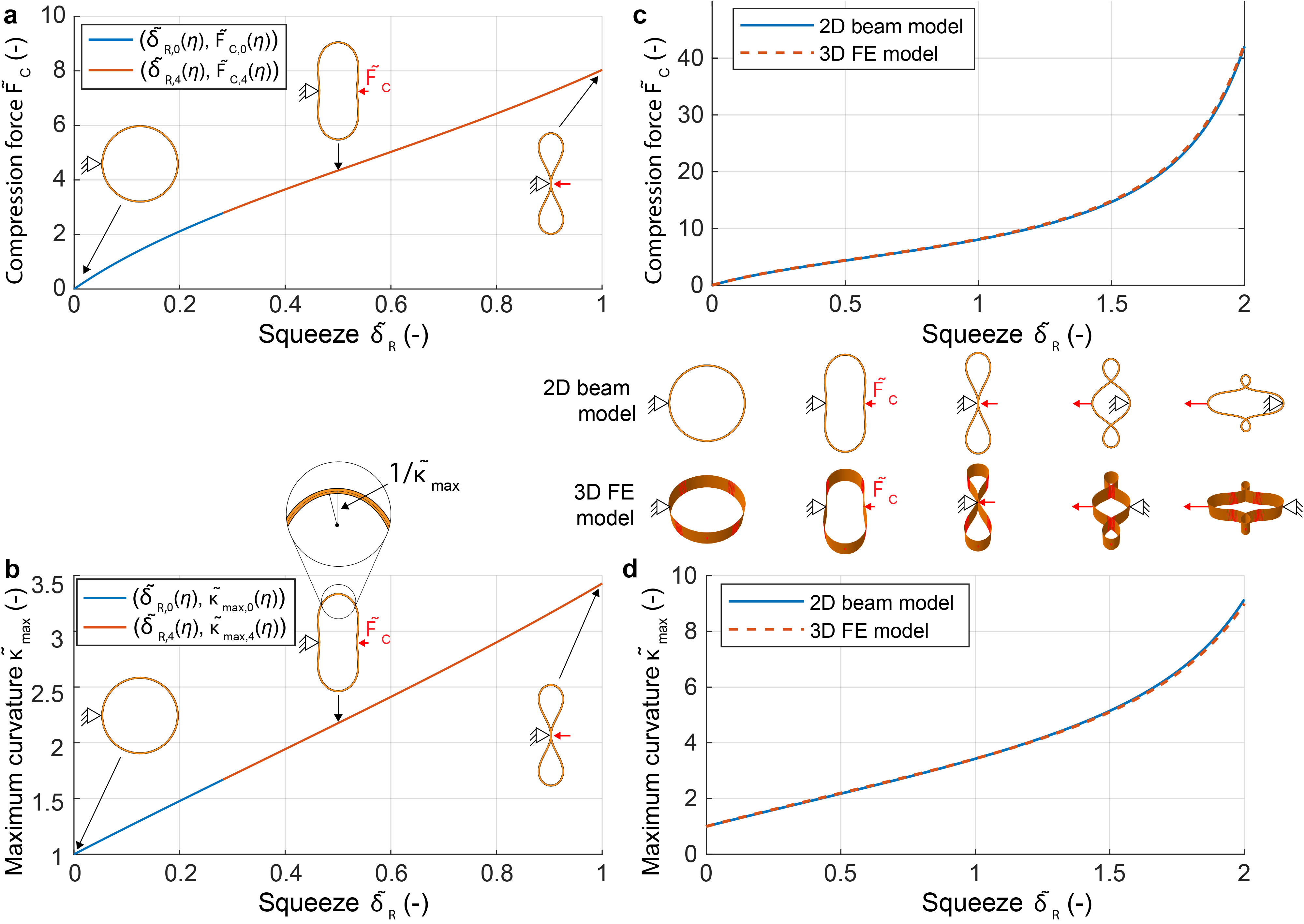}
  \caption{2D beam model and 3D FE validation. \textbf{a},\textbf{b}, Characteristic curves of the dimensionless compression force ($\tilde F_C$, \textbf{a}) and maximum curvature ($\tilde \kappa_{max}$, \textbf{b}) versus squeeze ($\tilde \delta_R$) with solutions for zero (blue) and four (red) inflection points for squeeze values from undeformed ($\tilde \delta_R$ = 0) to full squeeze ($\tilde \delta_R$ = 1). Characteristic curve coordinates are functions of the elliptic modulus $\eta$ and result from our 2D beam model. \textbf{c},\textbf{d}, Results of finite-element (FE) validation of the 2D beam model with a 3D FE model for $\tilde F_C$ versus $\tilde \delta_R$ (\textbf{c}) and $\tilde \kappa_{max}$ versus $\tilde \delta_R$ (\textbf{d}), extending the solution to self-intersecting values of squeeze up to $\tilde \delta_R$ = 2.}
  \label{fig:figure9}
\end{figure*}

The dimensionless force-compression characteristic, $\Gamma_F$, can be expressed as the union of two parametric force-compression curves: 
\begin{equation}
    \Gamma_F = \bigl(\tilde \delta_{R,0}(\eta),\tilde F_{C,0}(\eta) \bigr) \cup 
    \bigl(\tilde \delta_{R,4}(\eta),\tilde F_{C,4}(\eta) \bigr)
    \label{eq:gamma_F}
\end{equation}
where subscripts 0 and 4 denote the number of inflexion points of the deformed ring (\Crefsub{fig:figure9}{a}). The curves are defined as follows:
\begin{equation}
    \begin{split}
        \tilde \delta_{R,0}(\eta) = 
        1-\frac{\pi}{2}\biggl[1-\frac{2}{\eta^2} \biggl(1-\frac{\mathcal{E}\bigl(\frac{\pi}{4},\eta\bigr)}{\mathcal{K}\bigl(\frac{\pi}{4},\eta\bigr)}\biggr) \biggr],
        \hspace{2.5 cm} \eta\in[0,\sqrt{2}],\\
        \tilde F_{C,0}(\eta) = \frac{8}{\pi^2}\eta^2\mathcal{K}^2\bigl(\frac{\pi}{4},\eta\bigr),\hspace{4.9 cm} \eta\in[0,\sqrt{2}],\\
        \tilde \delta_{R,4}(\eta) = 
        1-\frac{\pi}{2}\biggl[2\frac{2\mathcal{E}(\eta)-\mathcal{E}\bigl(\mathrm{arcsin}\bigl(\frac{1}{\sqrt{2}\eta}\bigr),\eta\bigr)}{2\mathcal{K}(\eta)-\mathcal{K}\bigl(\mathrm{arcsin}\bigl(\frac{1}{\sqrt{2}\eta}\bigr),\eta\bigr)}-1 \biggr],
        \qquad \eta\in\biggl[\frac{1}{\sqrt{2}}, \bar \eta\biggr],\\
        \tilde F_{C,4}(\eta) = \frac{8}{\pi^2}\biggl[2\mathcal{K}(\eta)-\mathcal{K}\bigl(\mathrm{arcsin}\bigl(\frac{1}{\sqrt{2}\eta}\bigr),\eta\bigr)\biggr]^2,
        \hspace{1.75 cm} \eta\in\biggl[\frac{1}{\sqrt{2}}, \bar \eta\biggr]\\
    \end{split}
    \label{eq:curves_F}
\end{equation}
with $\mathcal{K}$ and $\mathcal{E}$ denoting incomplete elliptic integrals of the first and second kind, respectively~\cite{cazzolli2019snapping}, and $\bar \eta \approx$ 0.855, numerically estimated such that $\tilde \delta_{R,4}(\bar \eta)$ = 1.

The maximum non-dimensional curvature $\tilde\kappa_{max}$ = $2R\kappa_{max}$ always occurs at the midpoint of each beam. Similarly, the curvature-compression characteristic, $\Gamma_\kappa$, used to compute the maximum compression stress, is given by (\Crefsub{fig:figure9}{b}):
\begin{equation}
    \Gamma_\kappa = \bigl(\tilde \delta_{R,0}(\eta),\tilde\kappa_{max,0}(\eta) \bigr) \cup 
    \bigl(\tilde \delta_{R,4}(\eta),\tilde\kappa_{max,4}(\eta) \bigr)
    \label{eq:gamma_kappa}
\end{equation}
with:
\begin{equation}
    \begin{split}
        \tilde\kappa_{max,0}(\eta) = 
        \frac{2}{\eta}\sqrt{2\tilde F_{C,0}(\eta)},
        \hspace{1cm} \eta\in[0,\sqrt{2}],\\
        \tilde\kappa_{max,4}(\eta) = 
        2\eta\sqrt{2\tilde F_{C,4}(\eta)},
        \qquad \eta\in\biggl[\frac{1}{\sqrt{2}}, \bar \eta\biggr],\\
    \end{split}
    \label{eq:curves_kappa}
\end{equation}

The model assumes that strip thickness effects and non-uniformities caused by connection regions with rigid components are negligible. To validate the model and these assumptions, we used a non-dimensionalised version of the 3D simplified FE model developed for studying $k_\perp$ (see \Cref{sec:stiffness_def}). We conservatively assigned to the strip cross-section a width-to-thickness ratio of 10 (the lower bound we consider for the thin shell assumption to hold). We simulated squeezing by imposing kinematic boundary conditions at the left- and right-most airframe nodes. Reaction forces at these nodes determined $\tilde F_C$, while differentiation of discrete in-plane rotations at the mid-plane nodes determined $\tilde \kappa$. Even by comparing compression values $\tilde\delta_R >$ 1 (thus theoretically allowing self-intersection), the analytical model predicted the $\tilde F_C(\tilde \delta_R)$ and $\tilde \kappa_{max}(\tilde \delta_R)$ characteristics extracted from the FE model with normalized mean absolute errors nMAE$_F$ = nMAE$_\kappa$ = 0.32\% (\Crefmultisubfiglist{fig:figure9}{c}{d}). Both curves were uniformly resampled at $\Delta\tilde\delta_R$ = 0.002 increments using linear interpolation. nMAE values were normalized by $\tilde F_C(\tilde\delta_R )|_{\tilde\delta_R=2}$ = 42.48 and $\tilde \kappa_{max}(\tilde \delta_R)|_{\tilde\delta_R=2}$ = 8.99.

\subsection{Collision Model and Experiments}
\label{sec:collision_model}

We use peak acceleration at the control unit as the reference quantity for collision resilience because it directly relates to the shock absorption ratings provided in component datasheets. The normalisation factor of 50 g in \Cref{eq:res} derives from a standard threshold for shock and inertial tearing resistance of printed circuit boards and solder joints within circuit wiring, typically the most vulnerable points in control units.

To study collision resilience within FlexiQuad’s design space, we developed a simplified 2D parametric FE model using Abaqus 2024. The flexible laminate strips were modelled as a planar, ring-shaped beam using B21 elements. Battery housings and the control unit were modelled as planar parts with CPS4R shell elements and tied to the corresponding regions of the beam. The beam was assigned a rectangular cross-section representing the dimensions of the fibreglass strip. The shell components were assigned transverse thicknesses matching the physical heights of the attached hardware. A flat impact surface was represented by a narrow 2D rectangle composed of CPS4R shell elements, with fixed boundary conditions along the entire side opposite the point of collision. We simulated frontal collisions using an implicit dynamic analysis step, assigning the unconstrained FlexiQuad initial velocities of $v_0$ = 1, 3, and 5 m/s towards the fixed plane. The parametric FE model was systematically varied across a 16×16 log-log-spaced grid in the ($M$, $k$) design space (see \Cref{tab:table1}). Simulation solutions were obtained using Abaqus/Standard’s dynamic implicit solver using the analysis product default. From the simulation solutions, we extracted the time-dependent acceleration experienced by the CU and AUs, as well as the total reaction force at the impact plane. The peak magnitude of the acceleration at the CU determines $a_{CU,max} |_v$, used in the computation of the collision resilience metric $res$ with \Cref{eq:res}.

\begin{figure*}[htbp]
  \centering
  \includegraphics[width=0.99\textwidth]{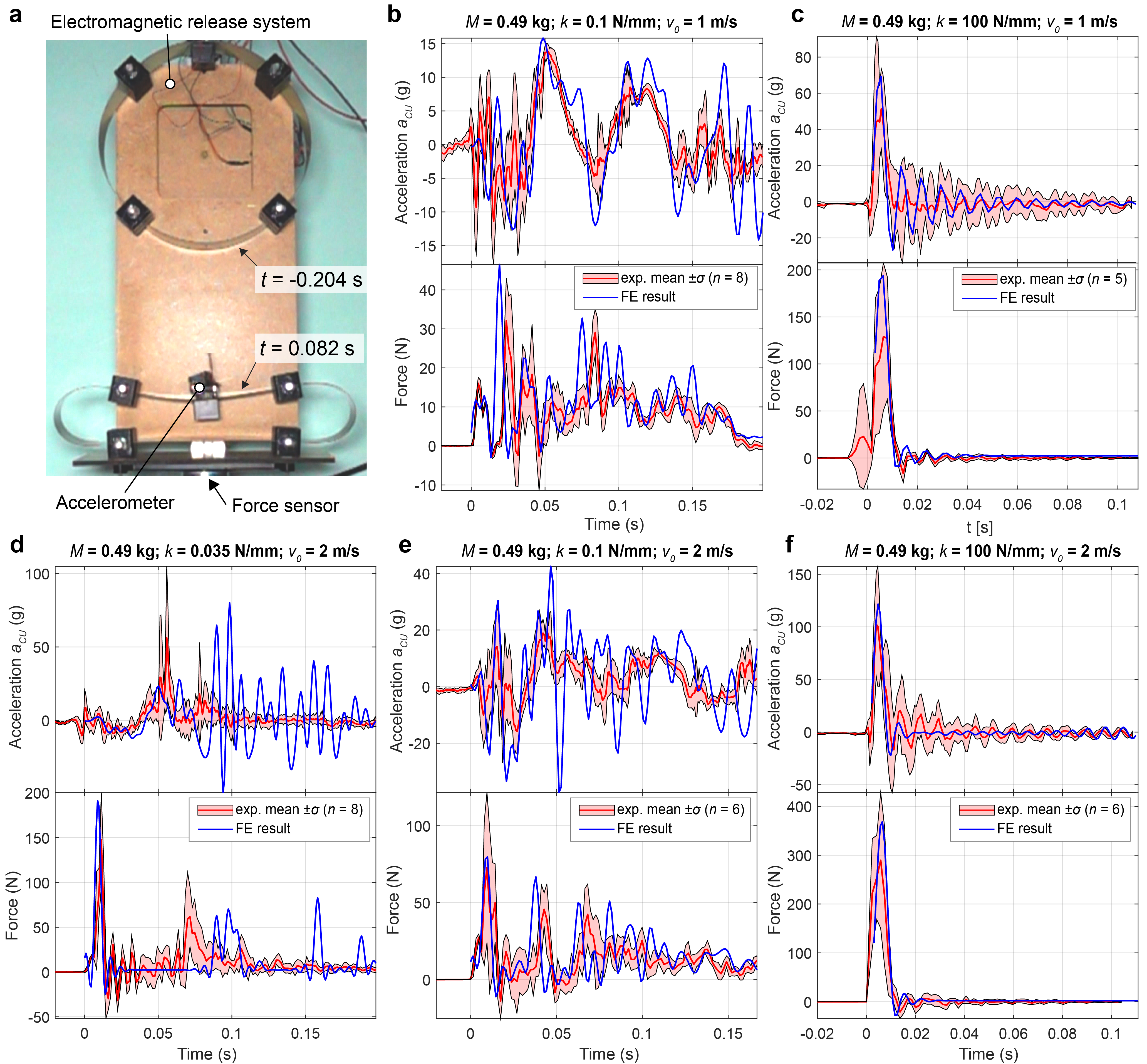}
  \caption{Experimental validation of FE analyses of frontal collisions. \textbf{a}, Snapshots from a drop-test ($M$ = 0.49, $k$ = 0.1, $v_0$ = 2 m/s). \textbf{b}-\textbf{f}, A selection of time profiles of impact-plane reaction force ($F$) and control unit acceleration ($a_{CU}$) at impact velocities $v_0$ = 1 and 2 m/s with mass ($M$) = 0.49 kg and varying stiffness ($k$) of the FlexiQuad dummy models. Finite-element FE analyses results (blue graphs) are compared with the mean curves from experimental measurements (red graphs), with indication of one (unbiased) standard deviation ($\pm \sigma$) from the mean (shaded red region). }
  \label{fig:figure10}
\end{figure*}

To validate the fidelity of the FE model, we conducted drop-tests with FlexiQuad mock-ups, measuring collision forces and CU acceleration (see \Crefsub{fig:figure10}{a}, and Supplementary Video~7). These mock-ups, constructed from the same fibreglass laminate used in FlexiQuad, featured battery housings filled with weights to match the intended FlexiQuad mass. We attached a 3-axis accelerometer (ADXL377), an Arduino Micro, and a micro-SD logger at the CU position, supplemented with additional weight to match the CU mass. The impact surface was a laser-cut, 5-mm-thick medium-density fibre-board (MDF) slab, attached via a 3D-printed bracket to a six-axis force-torque sensor (Rokubi, Bota Systems). We produced mock-up models corresponding to various ($M$, $k$) parameter combinations (listed in \Cref{tab:table2}) and performed frontal collision drop-tests at velocities of $v_0$ = 0.5, 1, 1.5, and 2 m/s, controlled through a height-adjustable, electromagnet-based release system. We ran FE collision simulations for the parameter combinations tested, adding the Earth’s gravity to match experimental conditions Supplementary Video~8). The simulated time evolution of the impact-plane reaction force ($F(t)$) and control unit acceleration ($a_{CU}(t)$) capture well the relevant peak timing and amplitude (see \Crefmultisubfigrange{fig:figure10}{b}{f}, and \Cref{tab:table2}). When rigid components collide with the plane, especially at faster velocities, stronger oscillatory behaviours are observed in the simulations’ accelerations after the first peak. These are due to unmodeled damping, but have limited effects on the relevance of the simulations within the scope of the analysis. Simulation accuracy was computed as the z-score ($z$) between the simulated peak and the normal distribution fit on the experimental peaks, both for $a_{CU}$ and $F$:
\begin{equation}
    z_q = \frac{\bigl(q_{peak,FE}-\mu_q \bigr)}{\sigma_q}, \qquad q = a_{CU}, F
    \label{eq:z_score}
\end{equation}
where $\mu_q$ is the mean and $\sigma_q$ the standard deviation (unbiased estimate) from experimental data. Most of the simulated peaks lie within three standard deviations from the experimental mean peaks.
\begin{table*}[htbp]
  \centering
  \small
  \setlength{\tabcolsep}{4pt}
  \begin{adjustbox}{max width=\textwidth}
  \begin{tabular}{|c|c|c|c|c|c|c|c|c|c|c|c|}
    \hline
    \thead{\textbf{Stiffness}\\$k$\\(N/mm)} &%
    \thead{\textbf{Velocity}\\$v_0$\\(m/s)} &%
    \thead{\textbf{Peak acc.}\\$a_{\text{peak}}^{(\text{FE})}$\\(g)} &%
    \thead{\textbf{Mean acc.}\\$\mu\!\left(a_{\text{peak}}^{(\text{FE})}\right)$\\(g)} &%
    \thead{\textbf{St. dev. acc.}\\$\sigma\!\left(a_{\text{peak}}^{(\text{FE})}\right)$\\(g)} &%
    \thead{$z_a$} &%
    \thead{$n_a$} &%
    \thead{\textbf{Peak force}\\$F_{\text{peak}}^{(\text{exp})}$\\(N)} &%
    \thead{\textbf{Mean force}\\$\mu\!\left(F_{\text{peak}}^{(\text{exp})}\right)$\\(N)} &%
    \thead{\textbf{St. dev. force}\\$\sigma\!\left(F_{\text{peak}}^{{(\text{exp})}}\right)$\\(N)} &%
    \thead{$z_F$} &%
    \thead{$n_F$} \\
    \hline\hline
    0.03 & 0.5 &  8.2 & 10.1 &  3.5 & \textbf{-0.55} & 8 &  36.3 &  32.6 &  7.4 &  \textbf{0.51} & 8 \\
    0.03 & 1.0 & 23.3 & 27.0 & 10.0 & \textbf{-0.37} & 7 &  49.3 & 107.2 & 28.6 & -2.02 & 7 \\
    0.03 & 1.5 & 59.1 & 74.0 & 23.4 & \textbf{-0.63} & 6 & 133.6 &  90.4 & 19.1 &  2.26 & 6 \\
    0.03 & 2.0 & 80.1 & 90.6 & 37.8 & \textbf{-0.28} & 8 & 191.3 & 167.4 & 22.6 &  1.06 & 8 \\
    0.10 & 0.5 &  9.7 &  8.8 &  1.3 &  \textbf{0.71} & 8 &  27.1 &  21.2 &  3.9 &  1.50 & 8 \\
    0.10 & 1.0 & 15.8 & 14.6 &  0.8 &  1.45 & 8 &  46.5 &  41.2 &  3.8 &  1.40 & 8 \\
    0.10 & 1.5 & 17.5 & 22.1 &  3.4 & -1.36 & 7 &  49.5 &  53.0 & 11.8 & \textbf{-0.30} & 8 \\
    0.10 & 2.0 & 42.5 & 28.3 &  6.9 &  2.08 & 8 &  79.7 & 113.4 & 20.4 & -1.65 & 6 \\
    0.42 & 0.5 & 11.6 &  9.6 &  1.4 &  1.41 & 6 &  22.5 &  24.4 &  1.6 & -1.19 & 7 \\
    0.42 & 1.0 & 23.9 & 19.1 &  2.4 &  2.01 & 8 &  37.3 &  47.1 &  8.1 & -1.21 & 8 \\
    0.42 & 1.5 & 34.7 & 28.1 &  3.2 &  2.08 & 8 &  43.9 &  64.1 & 14.5 & -1.39 & 7 \\
    0.42 & 2.0 & 53.2 & 48.7 &  3.5 &  1.28 & 8 &  95.1 &  86.7 & 10.0 &  \textbf{0.84} & 8 \\
    100  & 0.5 & 44.4 & 36.4 &  5.5 &  1.47 & 5 & 126.4 &  88.1 &  6.4 &  {\color{red}5.99} & 6 \\
    100  & 1.0 & 69.3 & 83.8 & 10.6 & -1.36 & 5 & 193.8 & 160.3 & 33.8 &  \textbf{0.99} & 6 \\
    100  & 1.5 & 90.4 & 110.3 &  6.6 & {\color{red}-3.01} & 6 & 255.3 & 224.9 & 41.8 &  \textbf{0.73} & 6 \\
    100  & 2.0 & 121.6 & 133.1 &  5.8 & -1.99 & 6 & 368.6 & 339.4 & 59.5 &  \textbf{0.49} & 6 \\
    \hline
  \end{tabular}
  \end{adjustbox}
  \caption{Drop test configurations. Selected values of stiffnesses ($k$) and collision velocities ($v_0$) used for the validation of the finite-element (FE) collision analyses with dummy FlexiQuad models of mass $M$ = 0.49 kg. Reported values of FE peak acceleration ($a_{\text{peak}}^{(\text{FE})}$) and force ($F_{\text{peak}}^{(\text{FE})}$), mean ($\mu(\cdot)$) and standard deviation ($\sigma(\cdot)$) of the experimental peak accelerations ($a_{\text{peak}}^{(\text{exp})}$) and forces ($F_{\text{peak}}^{(\text{exp})}$), with indication of z-scores ($z$) between FE and experimental values and number of experimental repetitions (n). Z-scores below 1 (in absolute value) in bold, those above 3 in red. }
  \label{tab:table2}
\end{table*}

The frontal collisions in \Crefsub{fig:figure3}{d} and Supplementary Video~3 were obtained by imposing remotely controlled body-angle commands to accelerate from hover to the desired collision velocity at a 1.5 m height in a straight line. Collisions were performed against a medium-density fibreboard (MDF) panel positioned vertically and measuring (2000$\times$1000$\times$25) mm. The trajectories’ kinematics, including collision velocity and acceleration, were tracked and logged at 300 Hz with a motion tracking system (OptiTrack, Qualisys). Collisions at $v_0$ = 3 and 4.5 m/s were tested.

We measured FlexiQuad’s resistance to glancing collisions by comparing the exchanged loads during a controlled collision at a propeller level with those experienced by an equivalent rigid multirotor in a bench-top testing setup (Supplementary Video~5). We tested a FlexiQuad model ($M$ = 405 g, $k$ = 0.1 N/mm) against a rigid quadrotor of identical mass, CU, and PUs, and featuring a rigid carbon-fibre frame (220-mm wheelbase, 4-mm thick). Both drone models were securely mounted at their CU positions to a 6-axis force-torque sensor (Bota Systems, Rokubi) via a 3D-printed bracket. The sensor was fixed to a stationary plate positioned beneath a swinging aluminium profile bar (0.96 kg, (1925$\times$20$\times$20) mm). An additional 0.044-kg mass was fixed at the bar’s free end to adjust the swing’s kinetic energy independently of collision velocity. The bar pivoted from one end, impacting the advancing propeller blade at a length of 110 mm from the bar’s free end. The bar's edge closest to the drone lay in a swinging plane offset 20 mm from the PU axis. An electromagnetic release mechanism initiated the swing from a height of 985 mm relative to its lowest point, ensuring a collision velocity of 5 m/s at the propeller level and a kinetic energy of 5.06 J, equivalent to that of the FlexiQuad model at 5 m/s.

\subsection{Agility Model and Experiments}
\label{sec:agility_model}

In the agility index definition (\Cref{eq:agt}), FlexiQuad’s peak acceleration, $\text{max}(\ddot\zeta_{FQ})$, corresponds to the highest value recorded across simulations increasing TWR from 2 to 8 in unit increments for each of the 256 ($M$, $k$) parameter combinations (\Cref{tab:table1}). The theoretical peak vertical acceleration for a given TWR is (TWR - 1) g, independent of body mass. As TWR = 8, the selected upper limit in this study, exceeds values typically used in racing quadrotors, a 7-g normalisation offers a conservative reference~\cite{foehn2022agilicious}. Accordingly, the linear agility index is given by:
\begin{equation}
    agt_z = \frac{\mathrm{max}(\ddot\zeta_{FQ})}{7}
    \label{eq:agt_function}
\end{equation}
Conversely, a rigid twin's angular agility varies with inertia, which scales with mass and size across the design space and, therefore, was computed for all 256 ($M$, $k$) combinations.

\begin{figure*}[htbp]
  \centering
  \includegraphics[width=0.99\textwidth]{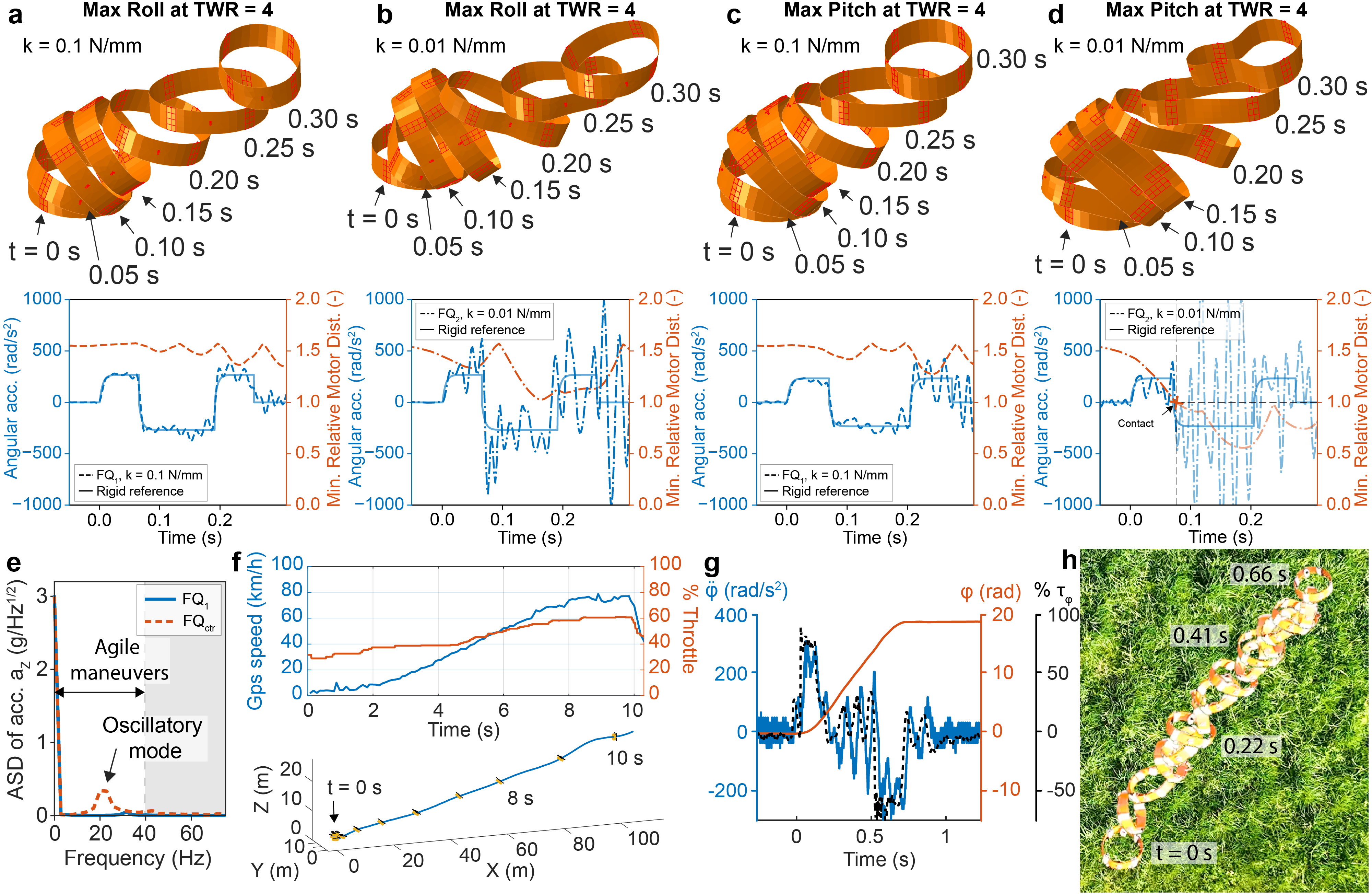}
  \caption{Additional agility and cruising speed results. \textbf{a}-\textbf{d}, Results from finite-element (FE) simulations of a dodging manoeuvre with maximal roll (\textbf{a}, \textbf{b}) and pitch (\textbf{c}, \textbf{d}) accelerations using FQ\textsubscript{1} ($M$ = 0.405 kg, $k$ = 0.1 N/mm, \textbf{a}, \textbf{c}) and FQ\textsubscript{2} ($M$ = 0.405 kg, $k$ = 0.01 N/mm, \textbf{b}, \textbf{d}) FlexiQuad models. Blue graphs compare angular acceleration about the studied axis versus time of a theoretical rigid twin (solid graph) with that of the FE-simulated FlexiQuad model (dashed and dash-dot graphs). Red graphs plot the minimum distance among all propulsion unit (PU) centres, normalised by the propeller diameter, extracted from FE results. \textbf{e}, Amplitude spectral density (ASD) versus frequency computed from the comparison of post-transient vertical accelerations ($a_z$, $0.05\leq $ time $ < 0.3$ s) in \Crefsub{fig:figure4}{h}. FlexiQuad model FQ\textsubscript{1} ($M$ = 0.405 kg, $k$ = 0.1 N/mm, blue) is compared against the counterpart of identical stiffness and concentrated battery masses FQ\textsubscript{ctr} (red). An oscillatory mode at $\sim$22 Hz \textendash\enspace only observed with FQ\textsubscript{ctr} \textendash\enspace is highlighted. The mode falls within the range of agile manoeuvres, wherein filtering control signals also slows down response, and hence, agility. \textbf{f}, Graphs of measured speed (blue) and commanded throttle percentage (red) versus time of a 10 s horizontal cruising speed test surpassing 75 km/h (top). Covered trajectory in 3D Cartesian coordinates from GPS measurements, with 10$\times$-magnified FlexiQuad icons showing the measured drone’s attitude every 1 s (bottom). \textbf{g},\textbf{h}, Graphs (\textbf{g}) and snapshots (\textbf{h}) of a four-flip roll trajectory with indication of measured roll acceleration ($\ddot \phi$ blue) and angle ($\phi$, red) and commanded percent reference roll torque (\% $\tau_\phi$, dashed black), using the FlexiQuad model FQ\textsubscript{1} ($M$ = 0.405 kg, $k$ = 0.1 N/mm) with TWR = 4.15.}
  \label{fig:figure11}
\end{figure*}

We developed a 3D finite-element model of FlexiQuad in Abaqus 2024 to study FlexiQuad’s agility. The airframe was modelled using S4R shell elements with rectangular cross-sections matching the fibreglass strips and assigned orthotropic lamina material properties. Rigid components, including batteries, PUs, and CU, were modelled as discrete point masses and inertias positioned at their respective centres of mass (COM), kinematically coupled to the airframe. Propeller thrust and drag torque were modelled as concentrated loads at the PU centres, with direction governed by local orientation (i.e. follow nodal rotation enabled in Abaqus). A global gravitational field in the -Z direction was applied. Desired force and torque commands followed first-order dynamics: $y(t)=\bigl(y_{des}-y(t)\bigr)/\tau$~\cite{faessler2017differential}, where $y(t)$ is the commanded variable, $y_{des}$ its target, and $\tau$ the time constant in seconds, conservatively set to $\tau$ = 0.001 s for $M <$ 0.1 kg, $\tau$ = 0.005 s for 0.1 kg $\leq M <$ 0.5 kg, $\tau$ = 0.01 s for 0.5 kg $\leq M <$ 1.5 kg, and $\tau$ = 0.02 s for $M \geq$ 1.5 kg. Simulations were performed with Abaqus/Standard’s dynamic implicit solver and no global boundary constraints, simulating free flight in all six degrees of freedom (DOFs). Simulated acceleration manoeuvres were categorised into vertical, roll, pitch, and yaw input commands. Linear agility was evaluated using a vertical acceleration step input. Angular agility was assessed through a three-phase ‘dodging’ manoeuvre, applying constant positive angular acceleration, then reversing to return to 0° with zero angular velocity (\Crefmultisubfigrange{fig:figure11}{a}{d}, and Supplementary Video~6). Step durations were precomputed assuming rigid-body dynamics to reach a maximum angle of 60°. Step amplitudes corresponded to those following from the selected TWR values from 2 to 8.

Mesh node kinematics were post-processed to compute COM accelerations. Linear accelerations were calculated as weighted averages of nodal accelerations. Angular accelerations were derived from COM kinematics, estimated by constructing a local frame: the Z-axis was defined as the normal to the best-fitting transverse plane; the X-axis as the vector from the CU to the foremost node; and the Y-axis as $u_y$ = $u_x \times u_z$. Angular rates were obtained by differentiating this frame and converting the result into Euler angle representation.
\begin{figure*}[htbp]
  \centering
  \includegraphics[width=0.99\textwidth]{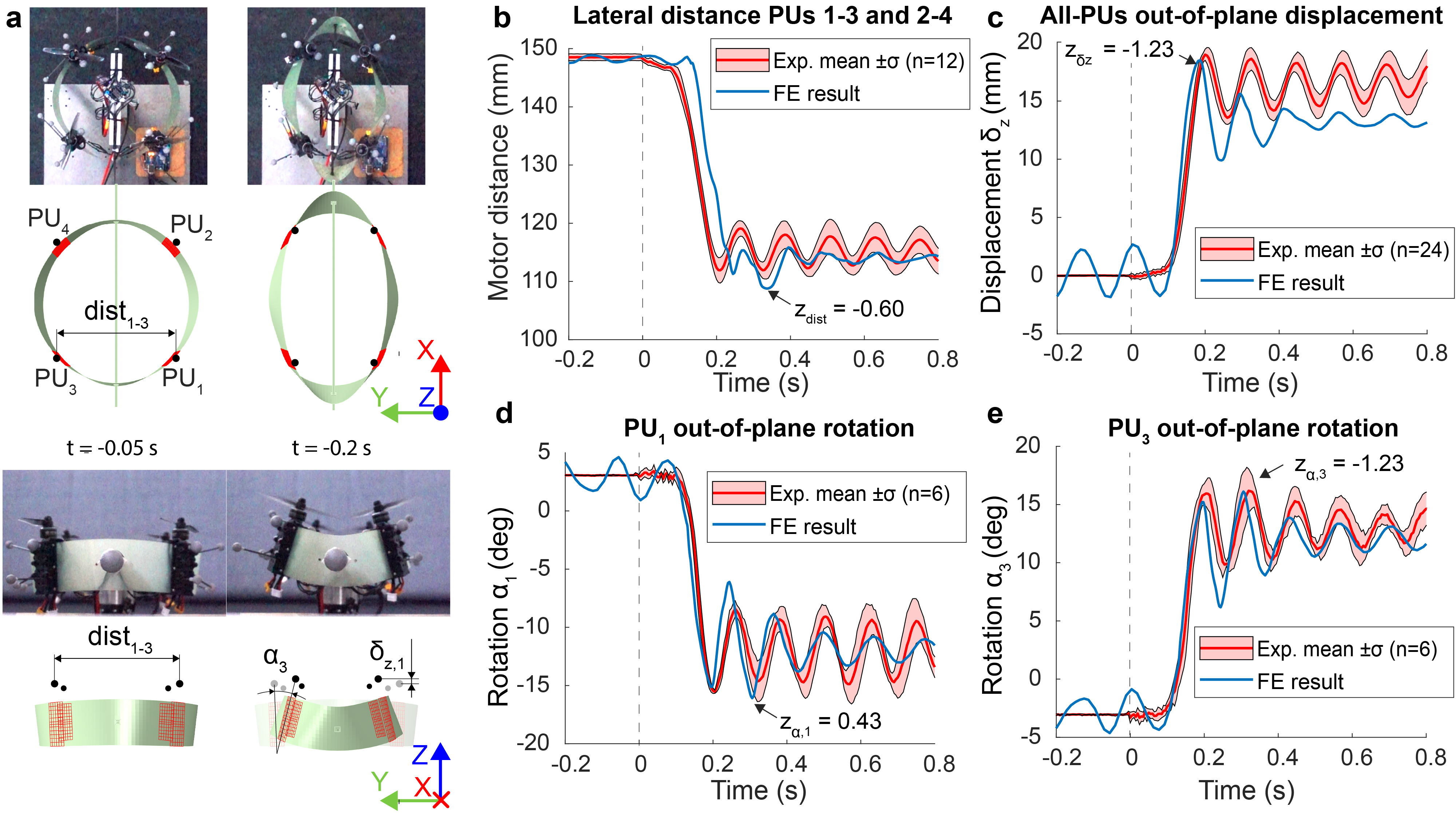}
  \caption{Experimental validation of FE analyses of dynamic acceleration commands. \textbf{a}, Top (top) and rear (bottom) views of bench-top experimental validation of finite-element (FE) simulations of maximum acceleration of a FlexiQuad model ($M$ = 0.375 kg, $k$ = 0.03 N/mm, TWR = 4.12), laterally and vertically constrained (Y and Z directions) by the rod passing through its longitudinal (X) axis. Comparison between experimental and FE results, both under gravity only (Time $t$ = -0.05 s, left) and at full throttle transient deformation ($t$ = 0.2 s, right). Indications of propulsion units (PUs), the distance between PU\textsubscript{1} and PU\textsubscript{3} ($dist_{1-3}$), and linear ($\delta_z$) and angular ($\alpha$) out-of-plane displacements are provided. \textbf{b}-\textbf{e}, Comparison graphs between FE results (blue) and experimental mean (solid red) $\pm$ standard deviation ($\sigma$, shaded red) of lateral distances $dist_{1-3}$ and $dist_{2-4}$ (\textbf{b}, n = 6 + 6, collectively pooled), out-of-plane displacements $\delta_z$ for all PUs (\textbf{c}, n = 4 $\times$ 6, pooled together), and out of plane rotations $\alpha$ at PU\textsubscript{1} (\textbf{d}, n = 6) and PU\textsubscript{3} (\textbf{e}, n = 6). Solid red shaded region indicating $\pm$ standard deviation, $\sigma$). All graphs report an indication of z-scores ($z$) calculated on graph peaks or valleys.}
  \label{fig:figure12}
\end{figure*}

For comparison with a centralised-battery design (\Crefmultisubfigrange{fig:figure4}{g}{j}), we modelled a twin configuration with identical $M$ and $k$, but two batteries concentrated along the roll axis, matching total battery mass and volume of FQ\textsubscript{1}. COM linear accelerations in \Crefsub{fig:figure4}{h} were calculated identically to the distributed configuration. Vertical angular deflections of the CU and AUs were quantified as the angle between these components’ local Z-axis directors and that of the quadrotor’s Z-axis, with AU deflections averaged across all four units (\Crefsub{fig:figure4}{j}):
\begin{equation}
    \begin{split}
        \alpha_{CU}(t) = \mathrm{cos}^{-1}\bigl( \mathbf{u}_{z,CU}(t)\cdot \mathbf{u}_z(t)  \bigr) \qquad\\
        \alpha_{AU}(t) = \frac{1}{4}\sum_{i=1}^4\mathrm{cos}^{-1}\bigl( \mathbf{u}_{z,AU_i}(t)\cdot \mathbf{u}_z(t)  \bigr)
    \end{split}
    \label{eq:angular_deflections}
\end{equation}
A frequency-domain comparison of COM accelerations revealed increased amplitude spectral density (ASD) centred around 20 Hz exclusively in the centralised model (\Crefsub{fig:figure11}{d}), which falls within the typical agile manoeuvre signal bandwidth for quadrotors.

To experimentally validate the FE model’s capability to approximate FlexiQuad’s dynamic deformation under impulsive flight loads, we performed bench tests on a constrained FlexiQuad model ($M$ = 0.375 kg, $k$ = 0.03 N/mm, TWR = 4.12). The drone was constrained laterally and vertically by a lubricated (silicone oil) carbon fibre rod inserted through the airframe centre, permitting translations and rotations about the longitudinal axis subject to finite friction (\Crefsub{fig:figure12}{a}). The tested configuration purposely featured a lower $k$ compared to prototype FQ\textsubscript{1}, thus allowing an evaluation with accentuated nonlinear deformations. Motor poses were tracked with spherical motion-capture markers during repeated 1-s full-throttle steps (n = 6). Corresponding simulation included detailed representations of the constraint rod and airframe-rod contact areas, with friction modelled using a penalty formulation (friction coefficient 0.02, see Supplementary Video~9). Simulated lateral distances between AUs, their out-of-plane deflections, and rotations closely matched experimental means, with peak transient differences falling within $\pm$1.25 standard deviations from experimental averages (z-scores ranged from -1.23 to 0.43; see \Crefmultisubfigrange{fig:figure12}{b}{e}).

Flight tests were conducted outdoors on a 405 g FlexiQuad prototype ($k$ = 0.1, TWR = 4.15). Vertical acceleration was measured from the ground using a manual full-throttle input for 1.1 s. Roll acceleration was tested from hover with a -100\% input for 0.33 s followed by +100\% for 0.49 s. For these tests, we added a GPS module tracking absolute position and velocity at a 10-Hz rate.

\section*{Supplementary Information}

Supplementary Videos 1\textendash9 will be publicly accessible after peer review. Early access can be provided on reasonable request.

\bibliographystyle{ieeetr}
\bibliography{references}

\end{document}